\documentclass{article}

\usepackage{arxiv}

\usepackage{graphicx}
\usepackage{amsmath}
\usepackage{amssymb}
\usepackage{booktabs}

\usepackage{subfigure}
\usepackage[subfigure]{tocloft}

\usepackage{soul}
\usepackage{mathtools}

\usepackage{multirow}
\usepackage{array}
\newcolumntype{H}{>{\setbox0=\hbox\bgroup}c<{\egroup}@{}}
\usepackage{makecell}

\usepackage{algorithm}
\usepackage{algorithmicx}
\usepackage{algpseudocode}

\usepackage{appendix}
\usepackage{comment}
\usepackage{caption}

\usepackage[T1]{fontenc}
\usepackage[utf8]{inputenc}
\usepackage{authblk}

\usepackage{bbding}

\usepackage{colortbl}
\usepackage{xcolor}
\definecolor{tabgray}{gray}{0.90}

\usepackage[pagebackref,breaklinks,colorlinks]{hyperref}

\usepackage[capitalize]{cleveref}
\crefname{section}{Sec.}{Secs.}
\Crefname{section}{Section}{Sections}
\Crefname{table}{Table}{Tables}
\crefname{table}{Tab.}{Tabs.}

\title{Multi-head Ensemble of Smoothed Classifiers for Certified Robustness}
\author[1]{Kun Fang}
\author[2]{Qinghua Tao}
\author[1]{Yingwen Wu}
\author[1]{Tao Li}
\author[1]{Xiaolin Huang}
\author[1]{Jie Yang}
\affil[1]{Department of Automation, Shanghai Jiao Tong University\\
{\tt\small
\{fanghenshao,yingwen\_wu,li.tao,xiaolinhuang,jieyang\}@sjtu.edu.cn}}

\affil[2]{School of Automation, Beijing Institute of Technology\\
{\tt\small qinghua.tao@bit.edu.cn}}

\date{}

\begin{document}
\maketitle
\begin{abstract}
Randomized Smoothing (RS) is a promising technique for certified robustness, and recently in RS the ensemble of multiple Deep Neural Networks (DNNs) has shown state-of-the-art performances due to its variance reduction effect over Gaussian noises.
However, such an ensemble brings heavy computation burdens in both training and certification, and yet under-exploits individual DNNs and their mutual effects, as the communication between these classifiers is commonly ignored in optimization.
In this work, we consider a novel \textit{ensemble}-based training way for a \textit{single} DNN with multiple augmented heads, named as SmOothed Multi-head Ensemble (SOME).
In SOME, similar to the pursuit of variance reduction via ensemble, an ensemble of multiple heads imposed with a cosine constraint inside a single DNN is employed with
much cheaper training and certification computation overloads in RS.
In such network structure, an associated training strategy is designed by introducing a circular communication flow among those augmented heads. 
That is, each head teaches its neighbor with the self-paced learning strategy using smoothed losses, which are specifically designed in relation to certified robustness.
The deployed multi-head structure and the circular-teaching scheme in SOME jointly contribute to the diversities among multiple heads and benefit their ensemble, leading to a competitively stronger certifiably-robust RS-based defense than ensembling multiple DNNs (effectiveness) at the cost of much less computational expenses (efficiency), verified by extensive experiments and discussions.
\end{abstract}

\section{Introduction}
\label{sec:intro}

Deep Neural Networks (DNNs) have been widely applied in various fields~\cite{deng2009imagenet,goodfellow2016deep}, but at the same time showed devastating vulnerability to adversarial samples.  
That is, images crafted with maliciously-designed perturbations, namely adversarial examples, can easily mislead well-trained DNNs into wrong predictions~\cite{szegedy2013intriguing,goodfellow2014explaining}.
To resist adversarial examples, there has been a series of empirical defenses against  adversarial attacks, \textit{e.g.}, adversarial training and its
variants~\cite{madry2018towards,zhang2019theoretically,rice2020overfitting}.
Besides, various methods for \textit{certified robustness}~\cite{wong2018provable,zhang2019towards,zhang2021boosting} have also attracted increasing attention in recent years, where researches focus on the robustness of
DNNs for images perturbed by Gaussian noises $\mathcal{N}(\mathbf{0},\sigma^2\boldsymbol{I})$. 
A certifiably-robust classifier guarantees that for any input $\mathbf{x}$, the predictions  are kept constant within its perturbed neighborhood  bounded commonly by the $\ell_2$  norm.

Randomized Smoothing (RS,~\cite{lecuyer2019certified,cohen2019certified,yang2020randomized}) is  considered as one of the most effective  $\ell_2$-norm certifiably-robust defenses.
With RS,  any base classifier can be formulated into a smoothed and certifiably-robust one by giving the most probable prediction over the Gaussian corruptions of the input.
~\cite{cohen2019certified} firstly proved a tight robustness guarantee of the smoothed classifier in RS and realized empirical implementation  to large networks and data sets, \textit{e.g.},
ResNet-50~\cite{he2016deep} on ImageNet~\cite{deng2009imagenet}.
Afterwards, various works have been developed for training a robust base classifier against Gaussian noises, so as to strengthen the certified robustness of the smoothed classifier in RS-based methods.
These works can be mainly categorized as two types: extra regularization terms~\cite{li2019certified,zhai2019macer,jeong2020consistency} and enhanced data augmentation
techniques~\cite{salman2019provably,jeong2021smoothmix}.

\begin{figure*}[t]
  \centering
   \includegraphics[width=0.95\linewidth]{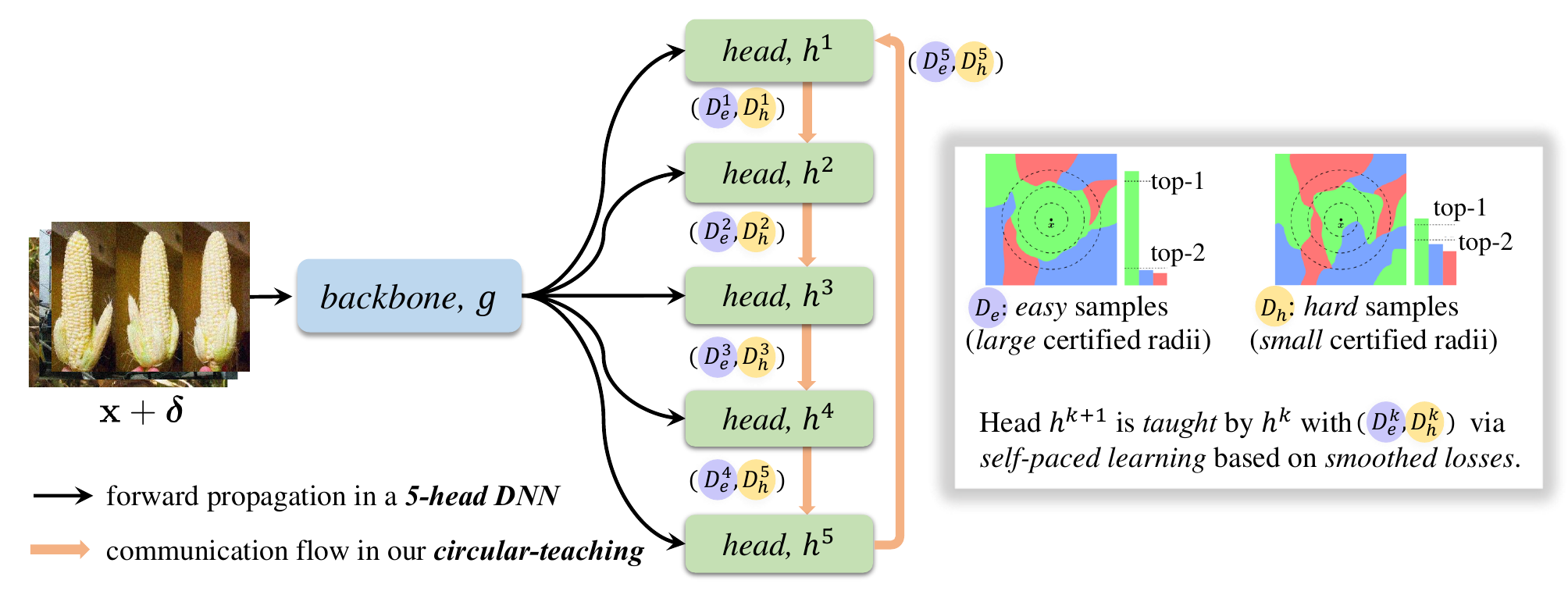}

   \caption{An illustration on the proposed SOME of a 5-head DNN. In our efficient structure for the ensemble, a common backbone $g$ is shared by the augmented 5 heads $h^1, \cdots, h^5$, which are trained to be mutually orthogonal to each other for diversified classifiers. 
   In the novel training with a circular communication flow between classifiers, the augmented head $h^{k+1}$ are optimized via the circular-teaching by its peer classifier $h^k$ with easy samples $D_e$ and hard ones $D_h$ in relation to the certified radii ($h^0\coloneqq h^5$). 
   The referred figures on the illustration of easy and hard samples are from \cite{horvath2021boosting}, and refer to \cref{sec:ct} for details on how to select easy and hard samples ($D_e$ and $D_h$) under this certified robustness task.
   }
   \label{fig:intro}
\end{figure*}

Recently, the \textit{ensemble} of multiple DNNs being the base classifier in RS has shown great potentials in certified robustness. 
The weighted logit ensemble of several DNNs trained from different random seeds~\cite{horvath2021boosting} or max-margin ensemble of fine-tuning multiple pretrained DNNs with extra constraints~\cite{yang2021certified}
have achieved significant improvements in higher approximated certified accuracy on the predictions of $\mathcal{N}(\mathbf{x},\sigma^2\boldsymbol{I})$. 
However, such an ensemble scheme demands considerably higher computation resources to train multiple large DNNs.
In particular, the resulting computational burden becomes more pronounced in the certification phase, because certifying each sample $\mathbf{x}$ can even require tens of thousands of samplings within
$\mathcal{N}(\mathbf{x},\sigma^2\boldsymbol{I})$~\cite{cohen2019certified}, leading to numerous inferences by forward propagation.
Consequently, certification time gets exacerbated approximately by the number of ensembled DNNs. 
Apart from the sacrifice of efficiency, these ensemble methods also ignore a straightforward and yet crucial fact that the current training of ensembled DNNs under-exploits the potentials of each individual DNN at hand and omits the mutual effects of each other in promoting the certified robustness.

Rather than the ensemble after the separate training of multiple DNNs, we in this paper consider a novel \textit{ensemble}-based training  for a \textit{single} DNN augmented with multiple heads. 
~\cite{horvath2021boosting} show that the ensemble of multiple DNNs trained separately can reduce the predictive variance against Gaussian noises, leading to certified robustness improvements.
Similar to the pursuit of variance reduction via ensemble, we employ an ensemble of multiple heads inside a single DNN with much cheaper training and certification computation overload.
Correspondingly, a novel training strategy has also been designed for the deployed multi-head DNN to fully explore the potentials for the certified robustness.
To be specific, a circular information communication flow among these heads is introduced into training, where each head \textit{teaches} its neighboring head with selected samples proceeding from ``easy'' to ``hard'' ones with specific considerations to certified robustness.
The proposed method is named as \textit{SmOothed Multi-head Ensemble (SOME)} with the workflow illustrated in  \Cref{fig:intro}. The contributions of SOME can be summarized as follows:

\begin{itemize}

    \item In the modeling, multiple heads are augmented in a single DNN, greatly alleviating computational burdens both in storage and time expenses required by multiple individual DNNs. A cosine constraint is constructed and imposed on the augmented heads, promoting diversity in the ensemble.
         
    \item The concept of communication  between ensembled heads is introduced in a joint optimization, enabling information exchange  to exploit potentials of individual heads and providing a novel perspective on the interaction in ensemble training.
    
  \item In the optimization, each head selects the mini-batch of samples  to teach its next neighboring head to update. The sample selection in such a circular-teaching scheme is conducted with self-paced
learning~\cite{kumar2010self,xu2015multi}, which proceeds the optimization from ``easy'' samples to ``hard'' ones, specifically defined in relation to the proposed \textit{smoothed loss} for certified robustness.
\end{itemize}

During certification, via the ensemble of heads, the proposed SOME shows competitive  certified robustness over the state-of-the-art (SOTA) ensemble method~\cite{horvath2021boosting} with much less computation overhead,
demonstrated by extensive experiments.
Note that SOME is compatible with all single-model-based methods, which can be plugged into training the heads in SOME via regularization~\cite{li2019certified,zhai2019macer,jeong2020consistency}  or data
augmentation~\cite{salman2019provably,jeong2021smoothmix}. 
SOME acts as an efficient and effective ensemble method, which successfully exemplifies a bridge connecting the methods of improving single models and integrating multiple models for certified robustness.

In the following,  \Cref{sec:related-work} outlines related works.
 \Cref{sec:methodology} elaborates SOME in terms of the multi-head structure and its  optimization.
 \Cref{sec:exp-results} presents extensive experiments on the certified robustness from different aspects.
 \Cref{sec:conclusion} discusses and concludes the proposed method.

%%%%%%%%%%%%%%%%%%%%%%%%%%%%%%%%%%%%%
%%%%%%%%%%%%%%%%%%%%%%%%%%%%%%%%%%%%%
%%%%%%%%%%%%%%%%%%%%%%%%%%%%%%%%%%%%%

\section{Related work}
\label{sec:related-work}

\subsection{Randomized smoothing}

Consider an arbitrary {\it base} classifier $f:\mathbb{R}^d\mapsto \mathbb{R}^C$ that takes the input $\mathbf{x}\in\mathbb{R}^d$ and gives $C$ logits \textit{w.r.t.} each class.
Then, denote $F(\mathbf{x})\coloneqq\arg\max_cf_c(\mathbf{x})$  as the function $\mathbb{R}^d\mapsto \mathcal C \triangleq \{1,\cdots,C\}$ predicting the  class of $\mathbf{x}$. 

In RS, a {\it smoothed} classifier $G$ is obtained as:
\begin{equation}
\label{eq:RS}
G(\mathbf{x})\coloneqq\mathop{\arg\max}\nolimits_{c}\mathcal{P}_{\boldsymbol{\delta}\sim\mathcal{N}({\bf 0},\sigma^2\boldsymbol{I})}\left(F(\mathbf{x}+\boldsymbol{\delta})=c\right).
\end{equation}
In \cref{eq:RS}, via Gaussian samplings within the neighborhood of $\mathbf{x}$, the smoothed classifier $G$  determines the top-1 and top-2 classes of $\mathbf{x}+\boldsymbol{\delta}$
with probabilities $p_A$ and $p_B$, respectively.
Based on the Neyman-Pearson lemma~\cite{neyman1933ix}, Cohen \textit{et al.}~\cite{cohen2019certified} proved a lower bound $R(G;\mathbf{x},y)$ of the $\ell_2$ certified radius of $G$ around $\mathbf{x}$ with true label $y$, such that for all $||\boldsymbol \delta||_2\leq R(G;\mathbf{x},y)$, the robustness of $G$, \textit{i.e.}, $G(\mathbf{x}+\boldsymbol \delta)=y$, is guaranteed in relation to  $p_A$ and  $p_B$~\cite{cohen2019certified}. 

Since the pioneering work on RS~\cite{lecuyer2019certified,cohen2019certified}, there have been various researches on training a more robust base classifier $f$.
The stability training~\cite{li2019certified} required similar predictions on both  original samples and noisy ones.
MACER~\cite{zhai2019macer} directly maximized the certified radius via a hinge loss.
In \cite{jeong2020consistency}, the predictive consistency was imposed within the input neighborhood.
These defenses~\cite{li2019certified,zhai2019macer,jeong2020consistency} can be  viewed as \textit{regularization-based} schemes, since additional terms are devised into the loss to train the base classifier.
Meanwhile, similar to the empirical defense of adversarial training~\cite{madry2018towards}, Salman \textit{et al.}~\cite{salman2019provably} and Jeong \textit{et al.}~\cite{jeong2021smoothmix} proposed to attack the smoothed classifier and simultaneously included the resulting adversarial examples in training the base classifier to enhance the certified robustness, which are regarded as \textit{data-augmentation-based} techniques.

The aforementioned works on RS enhance individual DNNs with a single classifier, while two concurrent works by Horv{\'a}th \textit{et al.}~\cite{horvath2021boosting} and Yang \textit{et al.}~\cite{yang2021certified} investigated how the \textit{ensemble} of multiple DNNs being the base classifier in RS improves the certified robustness.
In~\cite{horvath2021boosting}, some theoretical aspects were given that ensemble models boost the certified robustness of RS via variance reduction, and experiments also showed that the  average on the logits of multiple DNNs trained with different random seeds could achieve distinctive improvements.
In \cite{yang2021certified}, instead of the end-to-end training, it focused on a fine-tuning strategy for multiple pretrained DNNs, where  diversified gradients and large confidence margins were used for the fine-tuning. 
% The overhead computation in its pretraining for multiple DNNs is  already expensive, and the fine-tuning method in \cite{yang2021certified} further amplifies such computational expenses.
Despite of the improved certified accuracy, these ensemble methods greatly increase the computational burdens, \textit{i.e.},  amplifying both  storage and algorithmic complexities at least by the number of ensembled classifiers.
% Our work in contrast bridges methods using a single model and ensembling multiple models via an \textit{efficient} (cheaper training and certification) and \textit{effective} (stronger certified robustness) ensemble training strategy on an augmented multi-head structure to achieve a certifiably-robust single model.

Aside from those typical works on training a robust base classifier, RS has been extended to varied tasks. In \cite{chiang2020detection},
an RS-based certified defense for object detection against $\ell_2$-bounded attack was proposed.
Certifying the prediction confidence \cite{kumar2020certifying} was studied to achieve better certified guarantees.
Knowledge distillation was also introduced to accelerate the training in \cite{vaishnaviaccelerating}.
The watermarks hidden in DNNs could be well preserved via RS for privacy protection \cite{bansal2022certified}. 

\subsection{Co-teaching and self-paced learning}

Co-teaching was proposed in~\cite{han2018co} for combating label noise, and lots of variants \cite{yu2019does,nagarajan2024bayesian,chen2022compressing} were developed later.
Co-teaching involves two DNNs and requires that each network selects its small-loss samples to \textit{teach} (train) the peer network, as it assumes that the small-loss samples are more likely to be  the data with true labels. 
Therefore, two networks help each other select the  possibly clean samples to prevent overfitting the noisy labels in training. Co-teaching  is commonly suggested for  fine-tuning two well-trained DNNs.

Self-paced learning~\cite{kumar2010self,xu2015multi,he2023boosting,DBLP:conf/bmvc/ThangarasaT18} (SPL) has been proved as  a useful technique in avoiding bad local optima for better optimization results in machine learning.
In SPL, the model is trained with samples in an order of increasing difficulty, which facilitates the learning process. 
The ``difficulty'' of samples is defined in a similar way as how co-teaching thinks about the data with noisy labels: the smaller the loss, the easier the sample.  
SPL can be conducted by  hard weighting~\cite{kumar2010self} or soft weighting~\cite{xu2015multi}. 
The former conducts the optimization by only involving the  easiest samples within the mini-batch in each iteration, while the latter takes a sample-wisely weighted mini-batch during the iterative updates in training.

Note that although the proposed SOME does the teaching to its peer classifiers, it is essentially different from the existing co-teaching~\cite{han2018co}.
In our method,  multiple heads (classifiers) are designed to formulate a novel network structure on a single DNN for more economic and  effective ensembles. 
Our circular-teaching is intrinsically designed for allowing  communications among classifiers during the end-to-end training, so as to exploit more potentials of individual classifiers in the joint optimization for the ensemble. 
Our teaching mechanism is designed with the modified SPL in a specific relation to the proposed smoothed loss, not only for better optimization solutions but also for benefits of strengthening certified robustness.
More discussions on the related works can be found in \ref{app:related-works}.

\section{Methodology}
\label{sec:methodology}

In our proposed certified defense SOME, the multi-head structure (\cref{sec:mhead}) brings efficient ensemble prediction and certification, and the circular-teaching scheme (\cref{sec:ct}) regularizes the training of the multi-head DNN and leads to effective ensemble performance for improved certified robustness.
A table of notations is provided in \cref{tab:method-notations} for a convenient navigation on symbols involved in SOME.

\begin{table}[t]
  \caption{Table of notations involved in SOME.}
  \label{tab:method-notations}
  \centering
  \begin{tabular}{@{}cc@{}}
    \toprule

    Notations & Indications \\
    \midrule
    $h^k$ & The $k$-th head in the multi-head DNN \\
    $g$ & The backbone of the multi-head DNN \\
    $f^k$ & \makecell{The backbone followed by the $k$-th head, {\it i.e.}, $f^k\coloneqq h^k(g(\cdot))$} \\
    \midrule
    $({\bf x}_n,y_n)$ & The $n$-th sample ${\bf x}_n$ with its ground-truth label $y_n$ \\
    $\boldsymbol{\delta}_i$ & The $i$-th noise sampled from ${\cal N}({\bf0},\sigma^2\boldsymbol{I})$\\    $\sigma$ & The standard deviation of the Gaussian noise $\boldsymbol{\delta}_i$\\
    \midrule
    $\overline{\mathcal{L}^{k}_{\mathbf{x}_n}}$ & \makecell{The smoothed loss on the input sample $\mathbf{x}_n$ {\it w.r.t} the $k$-th head} \\
    $\nu_n^k$ & \makecell{The SPL coefficient on the input sample $\mathbf{x}_n$ {\it w.r.t} the $k$-th head} \\
    $\lambda$ & The threshold separating easy and hard samples \\
    \midrule
    $L$ & The number of heads in the multi-head DNN\\
    $m$ & The sampling times to determine the smoothed loss\\
    $N$ & The number of samples in a mini-batch of data\\
    \bottomrule
  \end{tabular}
\end{table}

\subsection{Augmented multi-head network}
\label{sec:mhead}

\noindent\textbf{Efficient structure.}
SOME deploys a single DNN augmented with multiple heads, each of which pertains a classifier for  certified robustness. 
This multi-head structure achieves the ensemble predictions using multiple classifiers from augmented heads. 
The ensemble is thereby realized in a much more efficient way and substantially mitigates the heavy computation resources required by the existing ensemble methods of integrating multiple individual DNNs~\cite{horvath2021boosting,yang2021certified}. 

The deployed  multi-head DNN consists of a backbone $g$, which is shared by $L$ augmented heads $h^1,\cdots,h^L$ with $h^i(g(\cdot)):\mathbb{R}^d\mapsto\mathbb{R}^C,i=1,\cdots,L$, as shown in \cref{fig:intro}. 
In prediction and certification, the model output is taken as the ensemble of these augmented  heads by averaging the logits from the classifiers:
\begin{equation}
\label{eq:ensemble}
f_{\mathrm{ens}}(\mathbf{x})=\frac{1}{L}\sum\nolimits_{k=1}^Lh^k(g(\mathbf{x})),
\end{equation}
where $f_{\mathrm{ens}}(\cdot)$ plays the role of the base classifier in RS without requiring extra modifications. 
Thus, the prediction and certification of the multi-head DNN in SOME can follow the typical functions \textsc{Predict} and \textsc{Certify} defined in~\cite{cohen2019certified} for predictions and certifications, respectively.

Besides, our forward propagation process differs from the existing ensemble methods \cite{horvath2021boosting,yang2021certified} in a more efficient way, because only additional computational overload in the augmented heads is required, rather than the complete cost in ensembling multiple DNNs.
Such forward propagation process of SOME directly contributes to a substantial reduction in the certification time compared with those ensemble methods computing multiple DNNs, since certifying each sample generally requires 100,000 samplings within the neighborhood of $\mathbf{x}$~\cite{cohen2019certified}.
\Cref{tab:method-cheap-structure} shows that over 60\%  certification time is reduced from 91.3s to 30.3s per sample, and more evidences on our efficiency can refer to \cref{sec:exp-runtime}, with detailed experiment settings and comparisons on FLOPs, training time, etc.

\begin{table}[t]
  \caption{Certification time of different methods. Our SOME distinctively improves the certification efficiency than the existing ensemble.}
  \label{tab:method-cheap-structure}
  \centering
  \begin{tabular}{@{}lccc@{}}
    \toprule

    Methods& Models & Per sample (s) & Total (h) \\
    \midrule
    Gaussian \cite{cohen2019certified} & 1 DNN & 17.3 & 2.41 \\
    Ensemble \cite{horvath2021boosting,yang2021certified} & 5 DNNs & 91.3 & 12.68 \\
    SOME (ours) & 5 heads & 30.3 & 4.20 \\
    \bottomrule
  \end{tabular}
\end{table}

\noindent\textbf{Diversifying parameters.} 
In ~\cite{horvath2021boosting}, the variance reduction via ensembling multiple DNNs trained from different random seeds distinctively improves the performance for certified robustness. 
Correspondingly, diversities between the multiple augmented heads in SOME are encouraged. 
Especially in such a structure with a common backbone,  the diversities across heads are  of particular interest and should be  designed delicately, avoiding extracting the same features and converging to identical heads.

A cosine constraint is thereby imposed on the parameters of all paired heads: 
\begin{equation}
\label{eq:ortho}
\mathcal{L}_{\rm cos}\coloneqq\sum_{i=1}^L\sum_{j=1,j\neq i}^L\frac{{\left \langle  h^i, h^j\right \rangle}^2}{|| h^i||\cdot||h^j||}.
\end{equation}
In \cref{eq:ortho}, ${\langle  h^i, h^j\rangle}$ is calculated as the inner product of the weights of $h^i$ and $h^j$, and $|| h^i||$ is determined as the $\ell_2$-norm of the weights of $h^i$.
This constraint promotes the diversity among the heads and explicitly guarantees that these heads do not converge to be identical with each other, indicating that the shared backbone is learned to simultaneously adapt to these diverse heads for pursuing good performance. 
Thus, in the considered network structure, a more adaptive backbone with diversified classifiers in augmented heads can be attained for the ensemble.
Similar constraints were also applied in other methods to boost diversities on classifiers for ensemble \cite{wortsman2021learning,fort2019deep}.
Ablation studies of $\mathcal{L}_{\rm cos}$ on the certified robustness can refer to \cref{app:ablation-cosine}.

\subsection{Optimization framework}
\label{sec:ct}
With the computation efficiency of the multi-head structure in SOME, we now introduce the optimization framework for SOME, aiming to achieve stronger and more efficient certified robustness than defenses of ensembling multiple DNNs.
Through the lens of peer-teaching classifiers, we propose a novel training strategy for SOME, circular-teaching, which allows communication and information exchange between the augmented heads.
The circular-teaching scheme is based on our Self-Paced Learning (SPL) that is specifically designed  for certified robustness, where the optimization of each head is taught by its neighboring head to progressively proceed from easy samples to hard samples selected by SPL.

\noindent\textbf{Sample easiness for certified robustness.}
Circular-teaching requires that the teaching among heads is executed differently on easy and hard samples.
In certified robustness, those samples with \textit{larger} certified radii are  \textit{easier} ones as they have better certified robustness, while those with \textit{smaller} certified radii are taken as the \textit{harder} ones.

As proven in \cite{cohen2019certified}, the certified radius of a sample $\mathbf{x}_n$ is determined by the probability gap between the top and runner-up predictions of the smoothed classifier. 
Nevertheless, it is nearly prohibitive to determining whether a sample is an easy or hard one by explicitly calculating the certified radius during the training phase.
Thereby, we adopt an efficient approximated proxy for the certified radius based on the theory results in \cite{cohen2019certified}, i.e., the following \textit{smoothed loss}:
\begin{equation}
\label{eq:sloss}
\overline{\mathcal{L}_{\mathbf{x}}}\coloneqq\frac{1}{m}\sum\nolimits_{i=1}^m\mathcal{L}_{\rm ce}(f(\mathbf{x}+\boldsymbol\delta_i),y).
\end{equation}
The smoothed loss $\overline{\mathcal{L}_{\mathbf{x}}}$ computes the average cross entropy loss of ${\bf x}_n$ within its neighborhood ${\cal N}({\bf x}_n,\sigma^2\boldsymbol{I})$ by sampling $m$ noises $\boldsymbol{\delta}_i\sim{\cal N}({\bf 0},\sigma^2\boldsymbol{I})$.
Thereby, for an easy(hard) sample with a large(small) certified radius, the smoothed classifier would have a high(low) probability of giving top predictions on noisy samples from the neighborhood \cite{cohen2019certified}, and produce a small(large) smoothed loss within the neighborhood.
Hence, the smoothed loss could serve as an appropriate proxy on the easiness of samples, which avoids explicitly calculating the certified radius.

\noindent \textbf{SPL for certified robustness.}
The naive SPL~\cite{kumar2010self} progressively includes samples into the optimization in a meaningful order, where only easy samples are used in the current iteration.  
SPL can also be extended with soft thresholds, that is, every sample is assigned with a coefficient in accordance with its easiness or hardness by the adapted logistic function~\cite{xu2015multi}, resulting in a sample-wisely weighted loss.
It has been proven theoretically and empirically that SPL helps avoid bad local optima and facilitate optimization~\cite{meng2017theoretical}. 
In this work, we adopt the soft thresholds in SPL and modify it for the circular-teaching in our multi-head network.

Given a sample $(\mathbf{x}_n,y_n)$ with its smoothed loss $\overline{\mathcal{L}_{\mathbf{x}_n}^k}$ {\it w.r.t} the $k$-th head $f^k\coloneqq h^k(g(\cdot))$, the SPL coefficient $\nu^k_n$ assigned to the sample $\mathbf{x}_n$ is calculated as
\begin{equation}
\label{eq:weights}
\nu^k_n=\left\{
    \begin{array}{ll}
         1 & \overline{\mathcal{L}^{k}_{\mathbf{x}_n}}\leq\lambda,\\
         \frac{1+e^{-\lambda}}{1+e^{\overline{\mathcal{L}^{k}_{\mathbf{x}_n}}-\lambda}} & \mathrm{otherwise}.
    \end{array}
    \right.
\end{equation}
In \cref{eq:weights}, samples with $\overline{\mathcal{L}^{k}_{\mathbf{x}_n}}\leq\lambda$ under $f^k$ are viewed as easy samples and are fully involved, \textit{i.e.}, $\nu^k_n=1$, while hard samples are weighted by the adapted logistic function~\cite{xu2015multi} modified based on the smoothed loss. 
In this way, the SPL coefficient $\nu^k_n$ serves as a sample-wise weight in the overall loss function, so that easy samples with $\nu^k_n=1$ and hard samples with $\nu^k_n<1$ contribute differently to the training of SOMEs.

\begin{algorithm}[t]
\caption{Training steps of SOME.}
\label{alg:some}
\begin{algorithmic}[1]
\Require A mini-batch of training samples ${\{(\mathbf{x}_n,y_n)\}}^N_{n=1}$, number of heads $L$,  sampling times $m$, threshold $\lambda$,  learning rate $\eta$, noise level $\sigma$. Note that $\theta$ denotes the to-be-optimized parameters of $g,h^1,...,h^L$.
\Ensure A deep neural network with augmented heads $f^k(\cdot)\coloneqq h^{k}(g(\cdot)),k\in\{1,2,...,L\}$.
\State Sample $\boldsymbol\delta_1,\cdots,\boldsymbol\delta_m$ from $\mathcal{N}({\bf0},\sigma^2\boldsymbol{I})$.
\State Determine smoothed loss $\overline{\mathcal{L}^k_{\mathbf{x}_n}}$ for each $f^k$ on $(\mathbf{x}_n,y_n)$.
\State Determine SPL coefficient $\nu^k_n$ as \cref{eq:weights} based on $\overline{\mathcal{L}^k_{\mathbf{x}_n}}$.
\State Determine the sample-weighted cross-entropy loss as $$\mathcal{L}_{\rm w}=\frac{1}{mNL}\sum_{k=1}^L\sum_{n=1}^N\sum_{i=1}^m{\color{red}\nu_n^{k-1}}\mathcal{L}_{\rm ce}({\color{red}f^k}(\mathbf{x}_n+\boldsymbol\delta_i),y_n).
$$
\State Determine $\mathcal{L}_{\mathrm{cos}}$ as \cref{eq:ortho}.
\State Update all $\theta$ simultaneously: $\theta\gets\theta-\eta\cdot\nabla_\theta(\mathcal{L}_{\rm w}+\mathcal{L}_{\rm cos})$.
\end{algorithmic}
\end{algorithm}

\noindent{\textbf{Circular-teaching in SOME.}}
As shown in \cref{fig:intro}, information circulates among heads by {peer-teaching} its next neighboring head with the selected samples.
To be specific, circular-teaching makes the $(k+1)$-th classifier $f^{k+1}$ to be trained by the sample mini-batch fed by its neighboring head $f^k$.
Hence, $f^{k+1}$ conducts  SPL in the view of $f^k$, and at the same time  $f^{k+1}$ analogously teaches its next neighboring head $f^{k+2}$, and so on, until the last head $f^L$ circulates the teaching to the first head $f^1$. 

At each training iteration, the $L$ SPL coefficients  $\nu_n^1,\cdots,\nu_n^L$ {\it w.r.t} $L$ heads are determined simultaneously on the sample easiness.
To flexibly implement the communication flow of the circular-teaching, one just needs to shift these $L$ coefficients circularly among heads, leading to the optimization objective:
\begin{equation}
\label{eq:opt}
\mathop{\min}_{g,h_1,\cdots,h_L}\frac{1}{mNL}\sum_{k=1}^L\sum_{n=1}^N\sum_{i=1}^m{\color{red}\nu_n^{k-1}}\mathcal{L}_{\rm ce}({\color{red}{f^k}}(\mathbf{x}_n+\boldsymbol\delta_i),y_n)+\mathcal{L}_{\rm cos}.
\end{equation}
Thus, there is no need to classify the samples nor exchange them between heads in computation, thanks to the flexible shift of coefficients.
In this way, the re-weighted cross entropy losses $\mathcal{L}_{\rm ce}$ of these heads are determined to update all the network parameters via stochastic gradient descent.
In \cref{alg:some}, key steps in each iterative update of SOME are outlined.

With the circular-teaching, each head provides its own perspective on the sample easiness and exchanges such information with peer heads.
In this way, each head can access more diversified information justified by its peer heads, and thereby all the heads are jointly trained for stronger diversities, which contributes to enhanced ensemble performance and promising certified robustness.

\begin{figure}[t]
  \centering
  \includegraphics[width=0.55\linewidth]{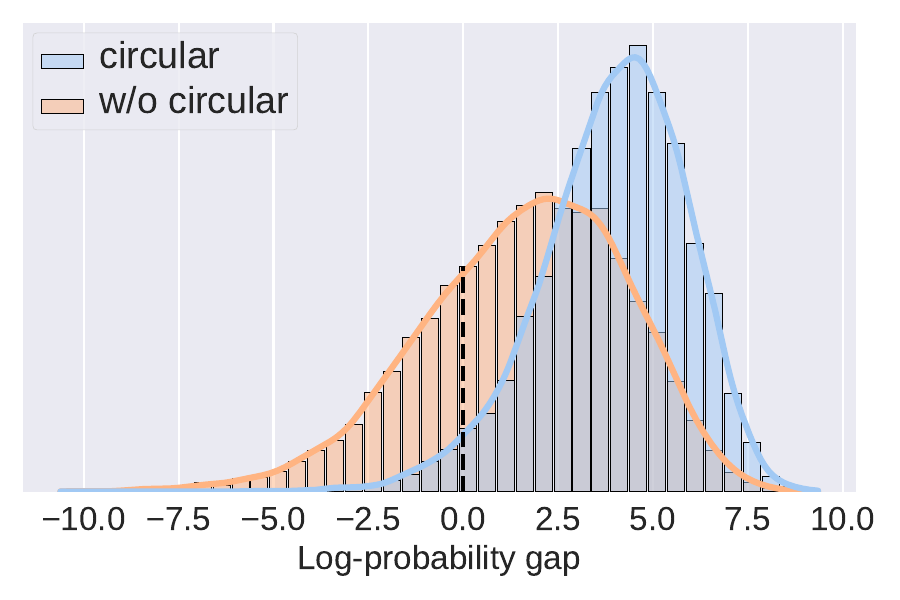}
  \caption{Comparisons of log-probability gap distributions of a randomly chosen test sample of CIFAR10, where 10,000 samplings are executed. In each histogram, the area in the left-side of the dashed line indicates misclassifications on noisy samples.}
  \label{fig:why-circular}
\end{figure}

We plot the log-probability distributions of two multi-head DNNs in \cref{fig:why-circular} for the cases with and without the circular information flow, respectively, showing  how such a communication flow helps  the predictions within $\mathcal{N}(\mathbf{x},\sigma^2\boldsymbol{I})$.
Given our ensemble prediction $f_{\rm ens}$  of the multi-head DNN in \cref{eq:ensemble},  the {log-probability gap}  \cite{jeong2020consistency} is calculated as $\log f^\Delta_y(\mathbf{x}+\boldsymbol\delta)-\mathop{\max}_{c\neq y}\log f^\Delta_c(\mathbf{x}+\boldsymbol\delta)$, where the probability outputs are denoted as $f^\Delta$ by using the softmax function on $f_{\rm ens}$.
Via numerous samplings of $\boldsymbol\delta$ from $\mathcal{N}(\mathbf{x},\sigma^2\boldsymbol{I})$, the distribution of log-probability gap  reflects the prediction variance of $f_{\rm ens}$ at $\mathbf{x}$ and the certified robustness of $G$ at $\mathbf{x}$. \Cref{fig:why-circular} presents a distinctively smaller area in the negative half on the horizontal-axis of the model with the circular communication, indicating that the circular order does bring a lower variance and stronger certified robustness.

\begin{table*}[!t]
  \caption{A comprehensive comparison among certified defenses on ACR and the approximated certified accuracy (\%) at different radii $r$ on CIFAR10 with noise levels $\sigma\in\{0.25,0.50,1.00\}$.
  Comparisons on the model capacities, including the number of DNNs, heads and the GFLOPs of different certified defenses, are also executed.
  Our results improving upon baseline methods are in bold and the overall best results are underlined. Results denoted by $\dagger$ are directly from~\cite{horvath2021boosting}.}
  \label{tab:comphensive-comparison-cifar10}
  \centering
  \resizebox{\textwidth}{!}{
  \begin{tabular}{@{}cl|ccc|ccccccccccc}
    \toprule
    \multirow{2}{*}{$\sigma$} & \multirow{2}{*}{Methods} & \multicolumn{3}{c|}{Model capacities} & \multirow{2}{*}{ACR $\uparrow$} & \multicolumn{10}{c}{Radius $r$}\\
    & & \# DNNs & \# heads & GFLOPs $\downarrow$ & & 0.00 & 0.25 & 0.50 & 0.75 & 1.00 & 1.25 & 1.50 & 1.75 & 2.00 & 2.25 \\
    \midrule
    \multirow{11}{*}{0.25} & Gaussian$^\dagger$~\cite{cohen2019certified} & 1 & 1 & 0.256 & 0.450 & 77.6 & 60.6 &  45.6 & 30.6 & 0.0 & 0.0 & 0.0 & 0.0 & 0.0 & 0.0 \\
    & Ensemble~\cite{horvath2021boosting} & \cellcolor{tabgray}5 & \cellcolor{tabgray}1 & \cellcolor{tabgray}1.280 & \cellcolor{tabgray}0.530 & \cellcolor{tabgray}82.2 & \cellcolor{tabgray}69.4 & \cellcolor{tabgray}53.8 & \cellcolor{tabgray}40.6 & 0.0 & 0.0 & 0.0 & 0.0 & 0.0 & 0.0 \\
    & \textbf{SOME} & \cellcolor{tabgray}1 & \cellcolor{tabgray}5 & \cellcolor{tabgray}{\bf0.594} & \cellcolor{tabgray}\textbf{0.537} & \cellcolor{tabgray}\underline{\textbf{83.4}} & \cellcolor{tabgray}\underline{\textbf{71.6}} & \cellcolor{tabgray}\textbf{54.4} & \cellcolor{tabgray}40.4 & 0.0 & 0.0 & 0.0 & 0.0 & 0.0 & 0.0  \\
    \cmidrule{2-16}
    & Consistency$^\dagger$~\cite{jeong2020consistency} & 1 & 1 & 0.256 & 0.546 & 75.6 & 65.8 & 57.2 & 46.4 & 0.0 & 0.0 & 0.0 & 0.0 & 0.0 & 0.0\\
    & Ensemble~\cite{horvath2021boosting} & \cellcolor{tabgray}5 & \cellcolor{tabgray}1 & \cellcolor{tabgray}1.280 & \cellcolor{tabgray}0.577 & \cellcolor{tabgray}75.6 & \cellcolor{tabgray}69.4 & \cellcolor{tabgray}60.0 & \cellcolor{tabgray}51.0 & 0.0 & 0.0 & 0.0 & 0.0 & 0.0 & 0.0\\
    & \textbf{SOME} & \cellcolor{tabgray}1 & \cellcolor{tabgray}5 & \cellcolor{tabgray}{\bf 0.594} & \cellcolor{tabgray}\textbf{0.580} & \cellcolor{tabgray}\textbf{77.6} & \cellcolor{tabgray}\textbf{69.4} & \cellcolor{tabgray}\textbf{61.4} & \cellcolor{tabgray}50.6 & 0.0 & 0.0 & 0.0 & 0.0 & 0.0 & 0.0\\
    \cmidrule{2-16}
    & SmoothMix~\cite{jeong2021smoothmix} & 1 & 1 & 0.256 & 0.545 & 77.6 & 68.4 & 56.6 & 44.2 & 0.0 & 0.0 & 0.0 & 0.0 & 0.0 & 0.0 \\
    & Ensemble~\cite{horvath2021boosting} & \cellcolor{tabgray}5 & \cellcolor{tabgray}1 & \cellcolor{tabgray}1.280 & \cellcolor{tabgray}0.585 & \cellcolor{tabgray}80.0 & \cellcolor{tabgray}70.4 & \cellcolor{tabgray}59.8 & \cellcolor{tabgray}50.4 & 0.0 & 0.0 & 0.0 & 0.0 & 0.0 & 0.0 \\
    & \textbf{SOME} & \cellcolor{tabgray}1 & \cellcolor{tabgray}5 & \cellcolor{tabgray}{\bf 0.594} & \cellcolor{tabgray}\underline{\textbf{0.589}} & \cellcolor{tabgray}77.8 & \cellcolor{tabgray}\textbf{70.4} & \cellcolor{tabgray}\underline{\textbf{62.6}} & \cellcolor{tabgray}\underline{\textbf{52.2}} & 0.0 & 0.0 & 0.0 & 0.0 & 0.0 & 0.0 \\
    \cmidrule{2-16}    
    & MACER$^\dagger$~\cite{zhai2019macer} & 1 & 1 & 0.256 & 0.518 & 77.4 & 69.0 & 52.6 & 39.4 & 0.0 & 0.0 & 0.0 & 0.0 & 0.0 & 0.0\\
    & SmoothAdv$^\dagger$~\cite{salman2019provably} & 1 & 1 & 0.256 & 0.527 & 70.4 & 62.8 & 54.2 & 48.0 & 0.0 & 0.0 & 0.0 & 0.0 & 0.0 & 0.0\\
    
    \midrule
    
    \multirow{12}{*}{0.50} & Gaussian$^\dagger$~\cite{cohen2019certified} & 1 & 1 & 0.256 & 0.535 & 65.8 & 54.2 & 42.2 & 32.4 & 22.0 & 14.8 & 10.8 & 6.6 & 0.0 & 0.0 \\
    & Ensemble~\cite{horvath2021boosting} & \cellcolor{tabgray}5 & \cellcolor{tabgray}1 & \cellcolor{tabgray}1.280 & \cellcolor{tabgray}0.634 & \cellcolor{tabgray}68.8 & \cellcolor{tabgray}60.6 & \cellcolor{tabgray}47.8 & \cellcolor{tabgray}39.2 & \cellcolor{tabgray}28.6 & \cellcolor{tabgray}20.0 & \cellcolor{tabgray}13.8 & \cellcolor{tabgray}8.4 & 0.0 & 0.0 \\
    & Ensemble$^\dagger$~\cite{horvath2021boosting} & \cellcolor{tabgray}10 & \cellcolor{tabgray}1 & \cellcolor{tabgray}2.560 & \cellcolor{tabgray}0.648 & \cellcolor{tabgray}69.0 & \cellcolor{tabgray}60.4 & \cellcolor{tabgray}49.8 & \cellcolor{tabgray}40.0 & \cellcolor{tabgray}29.8 & \cellcolor{tabgray}19.8 & \cellcolor{tabgray}15.0 & \cellcolor{tabgray}9.6 & 0.0 & 0.0 \\
    & \textbf{SOME} & \cellcolor{tabgray}1 & \cellcolor{tabgray}5 & \cellcolor{tabgray}{\bf 0.594} & \cellcolor{tabgray}\textbf{0.668} & \cellcolor{tabgray}\underline{\textbf{69.8}} & \cellcolor{tabgray}\underline{\textbf{61.8}} & \cellcolor{tabgray}\textbf{50.8} & \cellcolor{tabgray}\textbf{42.6} & \cellcolor{tabgray}\textbf{31.2} & \cellcolor{tabgray}\textbf{21.6} & \cellcolor{tabgray}\textbf{15.2} & \cellcolor{tabgray}\textbf{9.4} & 0.0 & 0.0 \\
    \cmidrule{2-16}
    & Consistency$^\dagger$~\cite{jeong2020consistency} & 1 & 1 & 0.256 & 0.708 & 63.2 & 54.8 & 48.8 & 42.0 & 36.0 & 29.8 & 22.4 & 16.4 & 0.0 & 0.0\\
    & Ensemble~\cite{horvath2021boosting} & \cellcolor{tabgray}5 & \cellcolor{tabgray}1 & \cellcolor{tabgray}{1.280} & \cellcolor{tabgray}0.741 & \cellcolor{tabgray}64.6 & \cellcolor{tabgray}57.2 & \cellcolor{tabgray}49.2 & \cellcolor{tabgray}44.4 & \cellcolor{tabgray}37.8 & \cellcolor{tabgray}31.6 & \cellcolor{tabgray}25.0 & \cellcolor{tabgray}18.6 & 0.0 & 0.0 \\
    & \textbf{SOME} & \cellcolor{tabgray}1 & \cellcolor{tabgray}5 & \cellcolor{tabgray}{\bf 0.594} & \cellcolor{tabgray}\textbf{0.745} & \cellcolor{tabgray}\textbf{65.2} & \cellcolor{tabgray}\textbf{59.6} & \cellcolor{tabgray}\underline{\textbf{52.4}} & \cellcolor{tabgray}\underline{\textbf{45.0}} & \cellcolor{tabgray}37.6 & \cellcolor{tabgray}30.4 & \cellcolor{tabgray}23.8 & \cellcolor{tabgray}17.6 & 0.0 & 0.0\\
    \cmidrule{2-16}
    & SmoothMix~\cite{jeong2021smoothmix} & 1 & 1 & {0.256} & 0.728 & 61.4 & 53.4 & 48.0 & 42.4 & 37.2 & 32.8 & 26.0 & 20.6 & 0.0 & 0.0\\
    & Ensemble~\cite{horvath2021boosting} & \cellcolor{tabgray}5 & \cellcolor{tabgray}1 & \cellcolor{tabgray}{1.280} & \cellcolor{tabgray}0.755 & \cellcolor{tabgray}59.8 & \cellcolor{tabgray}53.8 & \cellcolor{tabgray}48.2 & \cellcolor{tabgray}44.0 & \cellcolor{tabgray}39.4 & \cellcolor{tabgray}34.6 & \cellcolor{tabgray}28.0 & \cellcolor{tabgray}22.6 & 0.0 & 0.0 \\
    & \textbf{SOME} & \cellcolor{tabgray}1 & \cellcolor{tabgray}5 & \cellcolor{tabgray}{\bf 0.594} & \cellcolor{tabgray}\underline{\textbf{0.759}} & \cellcolor{tabgray}53.4 & \cellcolor{tabgray}50.2 & \cellcolor{tabgray}46.8 & \cellcolor{tabgray}\textbf{44.8} & \cellcolor{tabgray}\underline{\textbf{40.0}} & \cellcolor{tabgray}\underline{\textbf{35.8}} & \cellcolor{tabgray}\underline{\textbf{31.2}} & \cellcolor{tabgray}\underline{\textbf{25.6}} & 0.0 & 0.0 \\
    \cmidrule{2-16}
    & MACER$^\dagger$~\cite{zhai2019macer} & 1 & 1 & {0.256} & 0.668 & 62.4 & 54.4 & 48.2 & 40.2 & 33.2 & 26.8 & 19.8 & 13.0 & 0.0 & 0.0 \\
    & SmoothAdv$^\dagger$~\cite{salman2019provably} & 1 & 1 & {0.256} & 0.707 & 52.6 & 47.6 & 46.0 & 41.2 & 37.2 & 31.8 & 28.0 & 23.4 & 0.0 & 0.0 \\
    
    \midrule
    
    \multirow{13}{*}{1.00} & Gaussian$^\dagger$~\cite{cohen2019certified} & 1 & 1 & {0.256} & 0.532 & 48.0 & 40.0 & 34.4 & 26.6 & 22.0 & 17.2 & 13.8 & 11.0 & 9.0 & 5.8 \\
    & Ensemble~\cite{horvath2021boosting} & \cellcolor{tabgray}5 & \cellcolor{tabgray}1 & \cellcolor{tabgray}{1.280} & \cellcolor{tabgray}0.601 & \cellcolor{tabgray}49.0 & \cellcolor{tabgray}43.0 & \cellcolor{tabgray}36.4 & \cellcolor{tabgray}29.8 & \cellcolor{tabgray}24.4 & \cellcolor{tabgray}19.8 & \cellcolor{tabgray}16.4 & \cellcolor{tabgray}12.8 & \cellcolor{tabgray}11.2 & \cellcolor{tabgray}9.2 \\    
    & Ensemble$^\dagger$~\cite{horvath2021boosting} & \cellcolor{tabgray}10 & \cellcolor{tabgray}1 & \cellcolor{tabgray}{2.560} & \cellcolor{tabgray}0.607 & \cellcolor{tabgray}49.4 & \cellcolor{tabgray}44.0 & \cellcolor{tabgray}37.6 & \cellcolor{tabgray}29.6 & \cellcolor{tabgray}24.8 & \cellcolor{tabgray}20.0 & \cellcolor{tabgray}16.4 & \cellcolor{tabgray}13.6 & \cellcolor{tabgray}11.2 & \cellcolor{tabgray}9.4 \\
    & \textbf{SOME} & \cellcolor{tabgray}1 & \cellcolor{tabgray}5 & \cellcolor{tabgray}{\bf 0.594} & \cellcolor{tabgray}\textbf{0.656} & \cellcolor{tabgray}\underline{\textbf{50.6}} & \cellcolor{tabgray}\underline{\textbf{45.2}} & \cellcolor{tabgray}\textbf{37.8} & \cellcolor{tabgray}\textbf{31.6} & \cellcolor{tabgray}\textbf{27.8} & \cellcolor{tabgray}\textbf{22.4} & \cellcolor{tabgray}\textbf{18.4} & \cellcolor{tabgray}\textbf{14.8} & \cellcolor{tabgray}\textbf{12.0} & \cellcolor{tabgray}\textbf{10.6} \\
    \cmidrule{2-16}
    & Consistency$^\dagger$~\cite{jeong2020consistency} & 1 & 1 & {0.256} & 0.778 & 45.4 & 41.6 & 37.4 & 33.6 & 28.0 & 25.6 & 23.4 & 19.6 & 17.4 & 16.2 \\
    & Ensemble~\cite{horvath2021boosting} & \cellcolor{tabgray}5 & \cellcolor{tabgray}1 & \cellcolor{tabgray}{1.280} & \cellcolor{tabgray}0.809 & \cellcolor{tabgray}46.6 & \cellcolor{tabgray}42.0 & \cellcolor{tabgray}37.6 & \cellcolor{tabgray}33.0 & \cellcolor{tabgray}29.6 & \cellcolor{tabgray}25.8 & \cellcolor{tabgray}23.2 & \cellcolor{tabgray}21.0 & \cellcolor{tabgray}17.6 & \cellcolor{tabgray}16.2 \\
    & Ensemble$^\dagger$~\cite{horvath2021boosting} & \cellcolor{tabgray}10 & \cellcolor{tabgray}1 & \cellcolor{tabgray}{2.560} & \cellcolor{tabgray}0.809 & \cellcolor{tabgray}46.4 & \cellcolor{tabgray}42.6 & \cellcolor{tabgray}37.2 & \cellcolor{tabgray}33.0 & \cellcolor{tabgray}29.4 & \cellcolor{tabgray}25.6 & \cellcolor{tabgray}23.2 & \cellcolor{tabgray}21.0 & \cellcolor{tabgray}17.6 & \cellcolor{tabgray}16.2 \\
    & \textbf{SOME} & \cellcolor{tabgray}1 & \cellcolor{tabgray}5 & \cellcolor{tabgray}{\bf 0.594} & \cellcolor{tabgray}\textbf{0.830} & \cellcolor{tabgray}46.2 & \cellcolor{tabgray}\textbf{43.6} & \cellcolor{tabgray}\underline{\textbf{40.2}} & \cellcolor{tabgray}\underline{\textbf{35.8}} & \cellcolor{tabgray}\underline{\textbf{33.0}} & \cellcolor{tabgray}\textbf{28.2} & \cellcolor{tabgray}\textbf{25.6} & \cellcolor{tabgray}\textbf{22.0} & \cellcolor{tabgray}\textbf{18.8} & \cellcolor{tabgray}16.0 \\
    \cmidrule{2-15}
    & SmoothMix~\cite{jeong2021smoothmix} & 1 & 1 & {0.256} & 0.826 & 43.4 & 39.8 & 36.8 & 33.6 & 30.4 & 28.4 & 24.8 & 21.6 & 18.6 & 16.2\\
    & Ensemble~\cite{horvath2021boosting} & \cellcolor{tabgray}5 & \cellcolor{tabgray}1 & \cellcolor{tabgray}{1.280} & \cellcolor{tabgray}0.831 & \cellcolor{tabgray}43.8 & \cellcolor{tabgray}40.6 & \cellcolor{tabgray}38.2 & \cellcolor{tabgray}34.8 & \cellcolor{tabgray}31.2 & \cellcolor{tabgray}27.6 & \cellcolor{tabgray}24.0 & \cellcolor{tabgray}22.0 & \cellcolor{tabgray}19.2 & \cellcolor{tabgray}15.8 \\
    & \textbf{SOME} & \cellcolor{tabgray}1 & \cellcolor{tabgray}5 & \cellcolor{tabgray}{\bf 0.594} & \cellcolor{tabgray}\underline{\textbf{0.870}} & \cellcolor{tabgray}38.4 & \cellcolor{tabgray}36.2 & \cellcolor{tabgray}33.8 & \cellcolor{tabgray}32.0 & \cellcolor{tabgray}30.2 & \cellcolor{tabgray}\underline{\textbf{28.4}} & \cellcolor{tabgray}\underline{\textbf{26.2}} & \cellcolor{tabgray}\underline{\textbf{24.2}} & \cellcolor{tabgray}\underline{\textbf{22.0}} & \cellcolor{tabgray}\underline{\textbf{19.0}} \\
    \cmidrule{2-16}
    & MACER$^\dagger$~\cite{zhai2019macer} & 1 & 1 & {0.256} & 0.797 & 42.8 & 40.6 & 37.4 & 34.4 & 31.0 & 28.0 & 25.0 & 21.4 & 18.4 & 15.0 \\
    & SmoothAdv$^\dagger$~\cite{salman2019provably} & 1 & 1 & {0.256} & 0.844 & 45.4 & 41.0 & 38.0 & 34.8 & 32.2 & 28.4 & 25.0 & 22.4 & 19.4 & 16.6 \\
    \bottomrule
  \end{tabular}}
\end{table*}

\section{Numerical experiments}
\label{sec:exp-results}
Extensive experiments in this section are presented to evaluate SOME\footnote{ {\href{https://github.com/fanghenshaometeor/smoothed-multihead-ensemble}{https://github.com/fanghenshaometeor/smoothed-multihead-ensemble}.}} with comparisons to SOTA certified defenses, in terms of certified robustness (\cref{sec:exp-cifar10}), computational cost (\cref{sec:exp-runtime}), and ablation studies (\cref{sec:ablation-multi-head} and \cref{sec:ablation-thresholding}).

\begin{table*}[t]
  \caption{A comprehensive comparison among certified defenses on ACR and the approximated certified accuracy (\%) at different radii $r$ on ImageNet with noise levels $\sigma\in\{0.25,0.50,1.00\}$. Comparisons on the model capacities, including the number of DNNs, heads and the GFLOPs of different certified defenses, are also executed. Our results improving upon baseline methods are in bold and the overall best results are underlined. Results denoted by $\dagger$ are directly from~\cite{horvath2021boosting}.}
  \label{tab:comparison-imagenet}
  \centering
  \setlength{\tabcolsep}{1mm}
  % \resizebox{\textwidth}{!}{
  \begin{tabular}{cl|ccc|ccccccccc}
    \toprule
    \multirow{2}{*}{$\sigma$} & \multirow{2}{*}{Models} & \multicolumn{3}{c|}{Model capacities} & \multirow{2}{*}{ACR $\uparrow$} & \multicolumn{8}{c}{Radius $r$}\\
    & & \# DNNs & \# heads & GFLOPs $\downarrow$ & & 0.00 & 0.50 & 1.00 & 1.50 & 2.00 & 2.50 & 3.00 & 3.50 \\
    \midrule

    \multirow{8}{*}{0.25} & \multirow{3}{*}{Consistency$^\dagger$~\cite{jeong2020consistency}} & 1 & 1 & 4.122 & 0.512 & 63.0 & 54.0 & 0.0 & 0.0 & 0.0 & 0.0 & 0.0 & 0.0 \\
    & & 1 & 1 & 4.122 & 0.516 & 64.8 & 54.2 & 0.0 & 0.0 & 0.0 & 0.0 & 0.0 & 0.0 \\
    & & 1 & 1 & 4.122 & 0.509 & 64.4 & 53.8 & 0.0 & 0.0 & 0.0 & 0.0 & 0.0 & 0.0 \\
    \cmidrule{3-14}
    &Ensemble$^\dagger$~\cite{horvath2021boosting} & \cellcolor{tabgray}3 & \cellcolor{tabgray}1 & \cellcolor{tabgray}12.366 & \cellcolor{tabgray}0.545 & \cellcolor{tabgray}65.6 & \cellcolor{tabgray}57.0 & 0.0 & 0.0 & 0.0 & 0.0 & 0.0 & 0.0 \\
    % &\textbf{+ SOME} & 3 & \textbf{\underline{0.555}} & \textbf{66.8} & \underline{\textbf{57.8}} & 0.0 & 0.0 & 0.0 & 0.0 & 0.0 & 0.0 \\
    &{\bf SOME} & \cellcolor{tabgray}1 & \cellcolor{tabgray}3 &\cellcolor{tabgray}5.748 & \cellcolor{tabgray}\textbf{\underline{0.554}} & \cellcolor{tabgray}{\bf 66.0} & \cellcolor{tabgray}\underline{\textbf{57.2}} & 0.0 & 0.0 & 0.0 & 0.0 & 0.0 & 0.0 \\
    \cmidrule{2-14}
    & Gaussian$^\dagger$~\cite{cohen2019certified} & 1 & 1 & 4.122 & 0.477 & 66.7 & 49.4 & 0.0 & 0.0 & 0.0 & 0.0 & 0.0 & 0.0 \\
    & MACER$^\dagger$~\cite{zhai2019macer} & 1 & 1 & 4.122 & 0.544 & 68.0 & 57.0 & 0.0 & 0.0 & 0.0 & 0.0 & 0.0 & 0.0 \\
    & SmoothAdv$^\dagger$~\cite{salman2019provably} & 1 & 1 & 4.122 & 0.528 & 65.0 & 56.0 & 0.0 & 0.0 & 0.0 & 0.0 & 0.0 & 0.0 \\
    \midrule

    \multirow{9}{*}{0.50} & \multirow{3}{*}{Consistency$^\dagger$~\cite{jeong2020consistency}} & 1 & 1 & 4.122 & 0.793 & 56.0 & 48.0 & 39.6 & 34.0 & 0.0 & 0.0 & 0.0 & 0.0  \\
    & & 1 & 1 & 4.122 & 0.806 & 55.4 & 48.8 & 42.2 & 35.0 & 0.0 & 0.0 & 0.0 & 0.0 \\
    & & 1 & 1 & 4.122 & 0.799 & 54.0 & 48.0 & 41.2 & 35.2 & 0.0 & 0.0 & 0.0 & 0.0 \\
    \cmidrule{3-14}
    &Ensemble$^\dagger$~\cite{horvath2021boosting} & \cellcolor{tabgray}3 & \cellcolor{tabgray}1 & \cellcolor{tabgray}12.366 & \cellcolor{tabgray}0.868 & \cellcolor{tabgray}57.0 & \cellcolor{tabgray}52.0 & \cellcolor{tabgray}44.6 & \cellcolor{tabgray}38.4 & 0.0 & 0.0 & 0.0 & 0.0 \\
    % &+ \textbf{SOME} & 3 & \underline{\textbf{0.888}} & \underline{\textbf{59.4}} & \underline{\textbf{52.6}} & \underline{\textbf{45.8}} & \underline{\textbf{39.8}} & 0.0 & 0.0 & 0.0 & 0.0 \\
    &\textbf{SOME} & \cellcolor{tabgray}1 & \cellcolor{tabgray}3 &\cellcolor{tabgray}5.748 & \cellcolor{tabgray}\underline{\textbf{0.870}} & \cellcolor{tabgray}{\textbf{58.2}} & \cellcolor{tabgray}{51.2} & \cellcolor{tabgray}\underline{\textbf{45.2}} & \cellcolor{tabgray}\underline{\textbf{38.4}} & 0.0 & 0.0 & 0.0 & 0.0 \\
    \cmidrule{2-14}
    & Gaussian$^\dagger$~\cite{cohen2019certified} & 1 & 1 & 4.122 & 0.733 & 57.2 & 45.8 & 37.2 & 28.6 & 0.0 & 0.0 & 0.0 & 0.0 \\
    & MACER$^\dagger$~\cite{zhai2019macer} & 1 & 1 & 4.122 & 0.831 & 64.0 & 53.0 & 43.0 & 31.0 & 0.0 & 0.0 & 0.0 & 0.0 \\
    & SmoothAdv$^\dagger$~\cite{salman2019provably} & 1 & 1 & 4.122 & 0.815 & 54.0 & 49.0 & 43.0 & 37.0 & 0.0 & 0.0 & 0.0 & 0.0 \\ 
    & SmoothMix~\cite{jeong2021smoothmix} & 1 & 1 & 4.122 & 0.846 & 55.0 & 50.0 & 43.0 & 38.0 & 0.0 & 0.0 & 0.0 & 0.0 \\
    \midrule
    
    \multirow{5}{*}{1.00} & \multirow{3}{*}{Gaussian$^\dagger$~\cite{cohen2019certified}} & 1 & 1 & 4.122 & 0.856 & 42.4 & 37.2 & 31.4 & 25.8 & 19.6 & 15.4 & 12.0 & 8.4 \\
    & & 1 & 1 & 4.122 & 0.849 & 42.4 & 35.6 & 30.0 & 25.2 & 20.2 & 15.4 & 12.2 & 10.0 \\
    & & 1 & 1 & 4.122 & 0.839 & 40.8 & 35.6 & 29.4 & 25.8 & 20.4 & 15.6 & 12.2 & 8.0 \\
    \cmidrule{3-14}
    & Ensemble$^\dagger$~\cite{horvath2021boosting} & \cellcolor{tabgray}3 & \cellcolor{tabgray}1 &\cellcolor{tabgray}12.366 & \cellcolor{tabgray}0.968 & \cellcolor{tabgray}43.8 & \cellcolor{tabgray}38.4 & \cellcolor{tabgray}34.4 & \cellcolor{tabgray}29.8 & \cellcolor{tabgray}23.2 & \cellcolor{tabgray}18.2 & \cellcolor{tabgray}15.4 & \cellcolor{tabgray}11.4 \\
    % & \textbf{+ SOME} & 3 & \textbf{0.976} & \textbf{44.6} & \textbf{38.8} & 34.2 & \textbf{30.6} & \textbf{24.2} & 18.0 & 14.6 & \textbf{12.2} \\
    & \textbf{SOME} & \cellcolor{tabgray}1 & \cellcolor{tabgray}3 &\cellcolor{tabgray}5.748 & \cellcolor{tabgray}\textbf{1.002} & \cellcolor{tabgray}\textbf{46.8} & \cellcolor{tabgray}\textbf{40.8} & \cellcolor{tabgray}{\bf 35.4} & \cellcolor{tabgray}\textbf{30.4} & \cellcolor{tabgray}\textbf{23.2} & \cellcolor{tabgray}{\bf 19.4} & \cellcolor{tabgray}{\bf 15.4} & \cellcolor{tabgray}{11.0} \\
    % \cmidrule{2-14}
    % & Consistency$^\dagger$~\cite{jeong2020consistency} & - & - & - & 1.022 & 43.2 & 39.8 & 35.0 & 29.4 & 24.4 & 22.2 & 16.6 & 13.4 \\
    % & + Horv{\'a}th \textit{et al.}$^\dagger$~\cite{horvath2021boosting} & - & - & - & 1.108 & 44.6 & 40.2 & 37.2 & 34.0 & 28.6 & 23.2 & 20.2 & 16.4 \\
    % & \textbf{+ SOME} & - & - & - & \underline{\textbf{1.125}} & \textbf{45.4} & \textbf{40.8} & 37.0 & 33.8 & \underline{\textbf{29.2}} & \underline{\textbf{24.0}} & 19.6 & \underline{\textbf{17.2}} \\
    % \cmidrule{2-14}
    % & MACER$^\dagger$~\cite{zhai2019macer} & 1 & 1 & 4.122 & 1.008 & 48.0 & 43.0 & 36.0 & 30.0 & 25.0 & 18.0 & 14.0 & - \\
    % & SmoothAdv$^\dagger$~\cite{salman2019provably} & 1 & 1 & 4.122 & 1.011 & 40.6 & 38.6 & 33.8 & 29.8 & 25.6 & 20.6 & 18.0 & 14.4 \\
    % & SmoothMix~\cite{jeong2021smoothmix} & 1 & 1 & 4.122 & 1.047 & 40.0 & 37.0 & 34.0 & 30.0 & 26.0 & 24.0 & 20.0 & 17.0 \\
    \bottomrule
  \end{tabular} %}
\end{table*}

\noindent\textbf{Metrics.} Both the Average Certified Radius (ACR)~\cite{zhai2019macer} over the test set, and the approximated certified accuracy at different certified radii are adopted for evaluations.
Following the settings in ~\cite{cohen2019certified},  the Monte Carlo method is used  to estimate the certified radius of each test sample.
The approximated certified accuracy at a specific radius $r$ is calculated as the fraction of correctly-predicted samples with certified radii larger than $r$.

\noindent\textbf{Baseline methods.} For single-model-based methods, 3 representatives are involved, \textit{i.e.,} \underline{Gaussian}~\cite{cohen2019certified}: the pioneering work trained  on Gaussian corruptions of samples;  \underline{Consistency}~\cite{jeong2020consistency}: the SOTA regularization-based method by regularizing the prediction consistency within $\mathcal{N}(\mathbf{x},\sigma^2\boldsymbol{I})$; and  \underline{SmoothMix}~\cite{jeong2021smoothmix}: the SOTA data-augmentation-based method by mixing adversarial examples into training.
Besides, the current SOTA ensemble methods are considered, \textit{i.e.}, \underline{Ensemble}~\cite{horvath2021boosting}: the ensemble method via averaging the logits of multiple separately-trained DNNs.
Other popular RS-based methods are also included into  comparisons, including \underline{MACER}~\cite{zhai2019macer} and \underline{SmoothAdv}~\cite{salman2019provably}.
Details on how SOME augments Gaussian \cite{cohen2019certified}, Consistency \cite{jeong2020consistency} and SmoothMix \cite{jeong2021smoothmix} and the corresponding optimization objectives are provided in Appendix \ref{app:related-works}.

\begin{table*}
  \caption{Runtime comparisons on the CIFAR10 dataset. Our method greatly saves computational cost over the ensemble~\cite{horvath2021boosting}  and   distinctively improves ACR  over both single-model~\cite{cohen2019certified,jeong2020consistency} and ensemble~\cite{horvath2021boosting} baselines. Horv{\'a}th \textit{et al.}~\cite{horvath2021boosting} amplifies the running time of \cite{cohen2019certified,jeong2020consistency} by $\approx 5$. }
  \label{tab:exp-runtime}
  \centering
  \begin{tabular}{l|c|Hcc|cc|cc}
    \toprule
    \multirow{2}{*}{Methods} & \multirow{2}{*}{ACR} & \multirow{2}{*}{\# Params.} & \multirow{2}{*}{Models} & \multirow{2}{*}{GFLOPs} & \multicolumn{2}{c|}{Training time} & \multicolumn{2}{c}{Certification time} \\
    & & & & & per epoch (s) & total (h) & per sample (s) & total (h)\\
    \midrule
    Gaussian~\cite{cohen2019certified} & 0.532 & $1,730,714$ & 1 DNN & 0.256 & 39.6 & 1.65 & 17.3 & 2.41 \\
    \rowcolor{tabgray} Ensemble~\cite{horvath2021boosting} & 0.601 & $8,653,570$ & 5 DNNs & 1.280 & 198.0 & 8.25 & 91.3 & 12.68 \\
    \rowcolor{tabgray} SOME & {\bf 0.656} & $6,995,138$ & 5-head DNN & {\bf 0.594} & {\bf 155.5} & {\bf 6.48} & {\bf 30.3} & {\bf 4.20}  \\
    \midrule
    Consistency~\cite{jeong2020consistency} & 0.778 & $1,730,714$ & 1 DNN & 0.256 & 78.2 & 3.26 & 17.2 & 2.40 \\
    \rowcolor{tabgray} Ensemble~\cite{horvath2021boosting} & 0.809 & $8,653,570$ & 5 DNNs & 1.280 & 390.8 & 16.3 & 90.9 & 12.63 \\
    \rowcolor{tabgray} SOME & {\bf 0.830} & $6,995,138$ & 5-head DNN & {\bf 0.594} & {\bf 154.6} & {\bf 6.44} & {\bf 30.4} & {\bf 4.22} \\
    
    \bottomrule
  \end{tabular}
 \end{table*}

\noindent\textbf{Setups.} Following the common settings in certified robustness~\cite{cohen2019certified,jeong2020consistency,jeong2021smoothmix} for fair comparisons, the networks and data sets are used as ResNet-110~\cite{he2016deep} on CIFAR10 \cite{krizhevsky2009learning} and ResNet-50~\cite{he2016deep} on ImageNet \cite{deng2009imagenet} with noise levels varied on $\sigma\in\{0.25,0.5,1.0\}$.
The training and certification adopt the same noise level $\sigma$.
Appendix \ref{app:exp-cifar10} introduces more implementation details of models and hyper-parameters.

\subsection{Stronger certified robustness from SOME}
\label{sec:exp-cifar10}
SOME augments a single DNN with multiple heads trained via circular-teaching, and can be seamlessly applied to the existing methods based on single models. 
Following settings in the ensemble method~\cite{horvath2021boosting}, SOME augments the 3 representative single-model-based methods, Gaussian~\cite{cohen2019certified},  Consistency~\cite{jeong2020consistency}, and  SmoothMix~\cite{jeong2021smoothmix}, and gets compared with all the mentioned baseline models. 
The compared methods are evaluated based on the checkpoints released by {Horv{\'a}th \textit{et al.}}~\cite{horvath2021boosting}, and those results denoted by $\dagger$ are taken from the  paper~\cite{horvath2021boosting}.

\Cref{tab:comphensive-comparison-cifar10} illustrates new SOTA results of ACR and the approximated certified accuracy achieved by SOME on the CIFAR10 dataset. 
For the 3 methods using a single DNN, both the ensemble~\cite{horvath2021boosting} and our SOME substantially improve ACR and the approximated certified accuracy, showing the effectiveness of the ensemble for certified robustness. 
More specifically, compared to the defense ensembling 5 individual DNNs~\cite{horvath2021boosting}, our SOME using a single DNN with 5 heads further achieves stronger certified results, simultaneously with less than half the model capacities of~\cite{horvath2021boosting}, further verifying the advantageous effectiveness and efficiency of the proposed circular-teaching strategy.
Particularly in some cases, \textit{e.g.}, Gaussian baseline with $\sigma\in\{0.50,1.00\}$ and Consistency baseline with $\sigma=1.00$, SOME ensembling 5 heads in one single network even exceed the performance of ensembling up to 10 individual DNNs~\cite{horvath2021boosting}.

\Cref{tab:comparison-imagenet} shows the comparisons on the ImageNet dataset.
SOME with a 3-head DNN exhibits stronger certified robustness than the defense ensembling 3 DNNs~\cite{horvath2021boosting} with over half of the model capacities reduced. With all the 3 noise levels, SOME achieves the highest values of ACR and the approximated certified accuracy over other methods.
Detailed empirical settings are given in Appendix \ref{app:exp-imagenet}.

\subsection{More efficient computation from SOME}
\label{sec:exp-runtime}
In this section, the efficiency of SOME is comprehensively evaluated, where the single-model methods~\cite{cohen2019certified,jeong2020consistency} and the ensemble defense~\cite{horvath2021boosting} are involved for comparisons.
For a fair comparison,  each experiment in this section is executed on one single GPU of NVIDIA GeForce RTX 3060 (Ampere) with 12GB memory.
The compared single models Gaussian~\cite{cohen2019certified} and Consistency~\cite{jeong2020consistency} are based on ResNet-110 trained on CIFAR10 for 150 epochs with a noise level $\sigma=1.0$ and a batch size of 256.
The ensemble method~\cite{horvath2021boosting} trains 5 individual models with Gaussian~\cite{cohen2019certified} and Consistency~\cite{jeong2020consistency}, respectively, and then performs the ensemble. 
The proposed SOME is augmented with 5 heads on single models accordingly. The compared statistics mainly include the ACR, FLOPs, and time-consuming of training and certification phases.
Following previous works~\cite{cohen2019certified,horvath2021boosting},  we certify the CIFAR10 test set by evaluating every 20 samples.

\Cref{tab:exp-runtime} shows that the training and certification time of SOME gets significantly reduced. 
For both Gaussian~\cite{cohen2019certified} and Consistency~\cite{jeong2020consistency}, our novel ensemble reduces the certification time by $>50\%$ than the existing ensemble~\cite{horvath2021boosting}, and the training time is also greatly reduced, particularly in the case of Consistency~\cite{jeong2020consistency}.
The FLOPs also indicate a much cheaper algorithm complexity of SOME.
The efficiency advantage of such a multi-head structure in SOME can  be further pronounced by paralleling the forward propagation of all heads, which would be interesting in future implementations.

\subsection{Ablation studies on the essentials of SOME}
\label{sec:ablation-multi-head}

In this section, ablation studies on the essentials of SOME are executed, including the effect of the cosine constraint $\mathcal{L}_{\rm cos}$ (\cref{app:ablation-cosine}), the multi-head structure and the circular-teaching (\cref{app:ablation-mhead-ct}), different locations of heads (\cref{app:ablation-head-location}) and different numbers of heads (\cref{app:ablation-number-of-heads}).

\subsubsection{On the cosine constraint in SOME}
\label{app:ablation-cosine}

Ablation studies on the cosine constraint $\mathcal{L}_{\rm cos}$ are conducted to  investigate its effects.
The proposed $\mathcal{L}_{\rm cos}$ poses an orthogonality penalty on the multiple heads so as to obtain  more diversified classifiers to benefit the ensemble.
In this experiment, on CIFAR10 with a noise level $\sigma=1.0$, different SOME models are trained with and without $\mathcal{L}_{\rm cos}$, respectively. 
The corresponding ACR results of these models are listed in \cref{tab:ablation-ortho}, where 
the Consistency~\cite{jeong2020consistency} is employed as the base method.
Clearly, this constraint indeed  facilitates the performance, but the improvement is relatively more marginal, compared to the impacts of the keys in SOME, i.e., the leveraged multi-head structure and its circular-teaching mechanism, as shown in~\cref{tab:exp-ablation-q1-q2} later, which are the major contributions of this work.

\begin{table}[t]
    \caption{Ablation studies on the effectiveness of the cosine constraint $\mathcal{L}_{\rm cos}$.}
    \label{tab:ablation-ortho}
    \centering
    \begin{tabular}{c|ccc}
    \toprule
    ACR & $\sigma=0.25$ & $\sigma=0.50$ & $\sigma=1.00$ \\
    \midrule
     w/o $\mathcal{L}_{\rm cos}$ & 0.570 & 0.738 & 0.828\\
     w/ $\mathcal{L}_{\rm cos}$ & 0.580 & 0.745 & 0.830\\
    \bottomrule
    \end{tabular}
\end{table}

\begin{table*}[t]
  \caption{Ablation studies on the individual effectiveness of the multi-head structure (+ $L$-head) and the circular-teaching scheme of SOME (+ $L$-head + CT).}
  \label{tab:exp-ablation-q1-q2}
  \centering
  \begin{tabular}{@{}cl|ccccccccccc@{}}
    \toprule
    \multirow{2}{*}{$\sigma$} & \multirow{2}{*}{Methods} & \multirow{2}{*}{ACR} & \multicolumn{10}{c}{Radius $r$}\\
    & & & 0.00 & 0.25 & 0.50 & 0.75 & 1.00 & 1.25 & 1.50 & 1.75 & 2.00 & 2.25 \\
    \midrule
    \multirow{9}{*}{0.25} & Gaussian~\cite{cohen2019certified} & 0.450 & 77.6 & 60.6 &  45.6 & 30.6 & 0.0 & 0.0 & 0.0 & 0.0 & 0.0 & 0.0 \\
    & + 5-head & 0.508 & 80.8 & 67.6 & 52.0 & 37.2 & 0.0 & 0.0 & 0.0 & 0.0 & 0.0 & 0.0 \\
    & + 5-head + CT & 0.537 & 83.4 & 71.6 & 54.4 & 40.4 & 0.0 & 0.0 & 0.0 & 0.0 & 0.0 & 0.0 \\
    \cmidrule{2-13}
    & Consistency~\cite{jeong2020consistency} & 0.546 & 75.6 & 65.8 & 57.2 & 46.4 & 0.0 & 0.0 & 0.0 & 0.0 & 0.0 & 0.0 \\
    & + 5-head & 0.573 & 80.4 & 71.0 & 59.4 & 49.0 & 0.0 & 0.0 & 0.0 & 0.0 & 0.0 & 0.0 \\
    & + 5-head + CT & 0.580 & 77.6 & 69.4 & 61.4 & 50.6 & 0.0 & 0.0 & 0.0 & 0.0 & 0.0 & 0.0 \\
    \cmidrule{2-13}
    & SmoothMix~\cite{jeong2021smoothmix} & 0.545 & 77.6 & 68.4 & 56.6 & 44.2 & 0.0 & 0.0 & 0.0 & 0.0 & 0.0 & 0.0 \\
    & + 5-head & 0.588 & 80.0 & 72.2 & 60.2 & 51.2 & 0.0 & 0.0 & 0.0 & 0.0 & 0.0 & 0.0 \\    
    & + 5-head + CT & 0.589 & 77.8 & 70.4 & 62.6 & 52.2 & 0.0 & 0.0 & 0.0 & 0.0 & 0.0 & 0.0 \\
    
    \midrule
    \multirow{9}{*}{0.50} & Gaussian~\cite{cohen2019certified} & 0.535 & 65.8 & 54.2 & 42.2 & 32.4 & 22.0 & 14.8 & 10.8 & 6.6 & 0.0 & 0.0 \\
    & + 5-head & 0.629 & 68.0 & 59.8 & 49.6 & 38.8 & 30.2 & 19.4 & 13.8 & 8.0 & 0.0 & 0.0 \\
    & + 5-head + CT & 0.668 & 69.8 & 61.8 & 50.8 & 42.6 & 31.2 & 21.6 & 15.2 & 9.4 & 0.0 & 0.0 \\
    \cmidrule{2-13}
    & Consistency~\cite{jeong2020consistency} & 0.708 & 63.2 & 54.8 & 48.8 & 42.0 & 36.0 & 29.8 & 22.4 & 16.4 & 0.0 & 0.0 \\
    & + 5-head & 0.744 & 66.8 & 60.6 & 52.0 & 45.4 & 37.2 & 29.2 & 23.0 & 16.6 & 0.0 & 0.0 \\
    & + 5-head + CT & 0.745 & 65.2 & 59.6 & 52.4 & 45.0 & 37.6 & 30.4 & 23.8 & 17.6 & 0.0 & 0.0\\
    \cmidrule{2-13}
    & SmoothMix~\cite{jeong2021smoothmix} & 0.728 & 61.4 & 53.4 & 48.0 & 42.4 & 37.2 & 32.8 & 26.0 & 20.6 & 0.0 & 0.0\\
    & + 5-head & 0.750 & 59.0 & 53.6 & 48.2 & 42.6 & 38.8 & 34.0 & 29.2 & 23.0 & 0.0 & 0.0\\ 
    & + 5-head + CT & 0.759 & 53.4 & 50.2 & 46.8 & 44.8 & 40.0 & 35.8 & 31.2 & 25.6 & 0.0 & 0.0  \\

    \midrule
    \multirow{9}{*}{1.00} & Gaussian~\cite{cohen2019certified} & 0.532 & 48.0 & 40.0 & 34.4 & 26.6 & 22.0 & 17.2 & 13.8 & 11.0 & 9.0 & 5.8 \\
    & + 5-head & 0.590 & 49.4 & 44.2 & 37.0 & 29.0 & 23.0 & 19.2 & 16.2 & 13.2 & 10.8 & 8.0 \\
    & + 5-head + CT  & 0.656 & 50.6 & 45.2 & 37.8 & 31.6 & 27.8 & 22.4 & 18.4 & 14.8 & 12.0 & 10.6 \\
    \cmidrule{2-13}
    & Consistency~\cite{jeong2020consistency} & 0.778 & 45.4 & 41.6 & 37.4 & 33.6 & 28.0 & 25.6 & 23.4 & 19.6 & 17.4 & 16.2 \\
    & + 5-head & 0.774 & 47.2 & 43.4 & 38.6 & 34.4 & 29.4 & 25.8 & 22.6 & 19.2 & 16.8 & 14.6 \\
    & + 5-head + CT & 0.830 & 46.2 & 43.6 & 40.2 & 35.8 & 33.0 & 28.2 & 25.6 & 22.0 & 18.8 & 16.0 \\
    \cmidrule{2-13}
    & SmoothMix~\cite{jeong2021smoothmix} & 0.826 & 43.4 & 39.8 & 36.8 & 33.6 & 30.4 & 28.4 & 24.8 & 21.6 & 18.6 & 16.2 \\
    & + 5-head & 0.859 & 44.0 & 41.2 & 37.8 & 34.8 & 31.4 & 28.2 & 25.6 & 22.2 & 20.2 & 17.4 \\ 
    & + 5-head + CT & 0.870 & 38.4 & 36.2 & 33.8 & 32.0 & 30.2 & 28.4 & 26.2 & 24.2 & 22.0 & 19.0\\
    \bottomrule
  \end{tabular}
\end{table*}

\subsubsection{On the multi-head structure and the circular-teaching}
\label{app:ablation-mhead-ct}
The individual effects of the multi-head structure and the circular-teaching are investigated in ablation studies.
\Cref{tab:exp-ablation-q1-q2} shows that simply augmenting an individual model (``+ 5-head'') without circular-teaching can improve the certified robustness of the 3 base methods in ACR and the approximated certified accuracy.
It indicates that the ensemble of multiple heads inside one single DNN helps variance reduction on the ensemble predictions within $\mathcal{N}(\mathbf{x},\sigma^2\boldsymbol{I})$.
On top of the multi-head structure, the proposed circular-teaching scheme further enhances the certified robustness, reflected by the highest values of ACR and the approximated certified accuracy of ``+ 5-head + CT''.
It verifies that circular-teaching scheme further enhances the diversity among heads via the circular communication and the information exchange, which promotes the ensemble performance and boosts the certified robustness.

\subsubsection{On the different head locations}
\label{app:ablation-head-location}

\begin{figure}[t]
  \centering
   \includegraphics[width=0.8\linewidth]{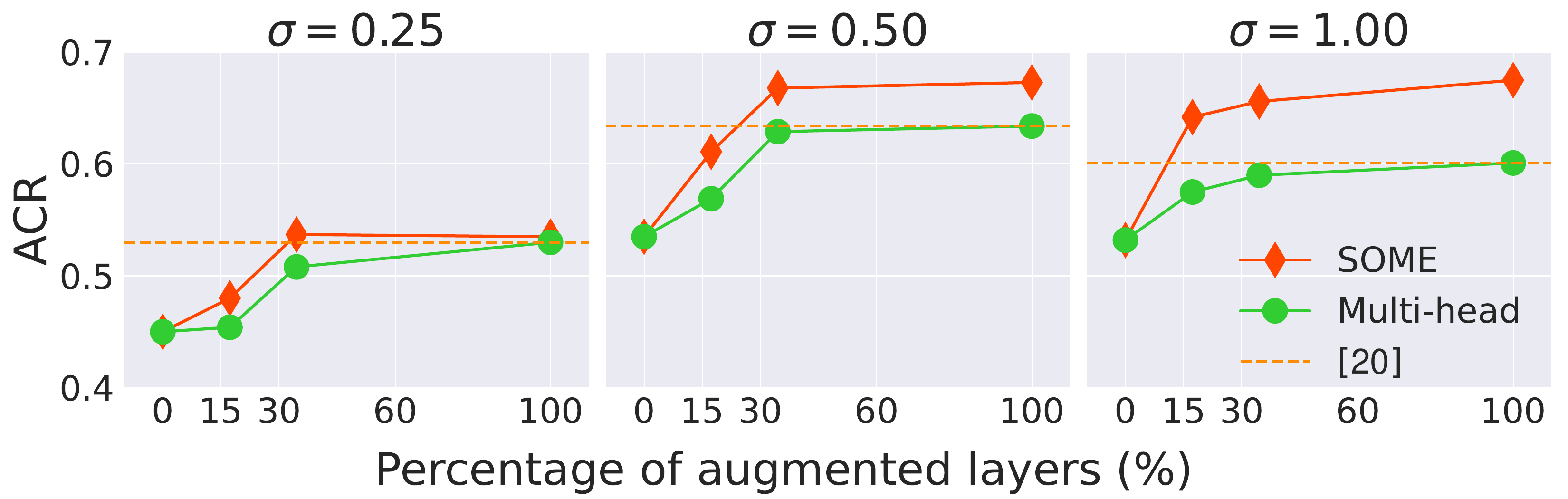}
   \caption{Ablation studies on where to augment heads.}
   \label{fig:ablation-where-to-arg}
\end{figure}

We further explore the influence of different locations where heads get augmented.
Generally in this paper, the certain location to augment heads of SOME is placed right after the 2nd residual block, resulting in  $\approx34\%$  layers being augmented in each head, and see details in \cref{fig:app-arch}.
In this experiment, we choose one more different augmenting position for ResNet-110 on CIFAR10, i.e., the middle of the 3rd residual block, where each head accounts for $\approx17\%$ layers in the 110-layer residual network.
\Cref{fig:ablation-where-to-arg} shows the ACR results {\it w.r.t} different augmentation locations in ResNet-110 with noise levels $\sigma\in\{0.25,0.50,1.00\}$, where 
the base method Gaussian~\cite{cohen2019certified} for training is employed.
We consider two cases: a multi-head DNN trained with and without circular-teaching, denoted as ``SOME'' and ``Multi-head'' in \cref{fig:ablation-where-to-arg}, respectively.
The horizontal axis indicates the percentage of layers in one head over total layers: 0\%,17\%,34\%, and 100\%. The
``0\%'' implies a standard DNN, and ``100\%''  indicates multiple individual DNNs.

As shown in \cref{fig:ablation-where-to-arg}, as the number of augmented layers increases, the ACR of both ``SOME'' and ``Multi-head'' improves a lot over one standard DNN.
Besides, in all the 3 different augmentation proportions (17\%, 34\% and 100\%), SOME all shows consistent superiority over the naive ensemble of multiple heads, indicating that the communication among heads benefits diversities of heads and thus boosts certified robustness.
Specifically, \cref{fig:ablation-where-to-arg} also illustrates that by only augmenting nearly or less than 30\% of total layers with appropriate carefully-designed communication mechanisms, one multi-head DNN can possibly outperform the ensemble of multiple individual DNNs~\cite{horvath2021boosting} (shown by the orange dashed lines) in certified robustness, further verifying our SOME as an efficient and effective ensemble method for certified robustness.

\begin{figure}[t]
  \centering
   \includegraphics[width=0.8\linewidth]{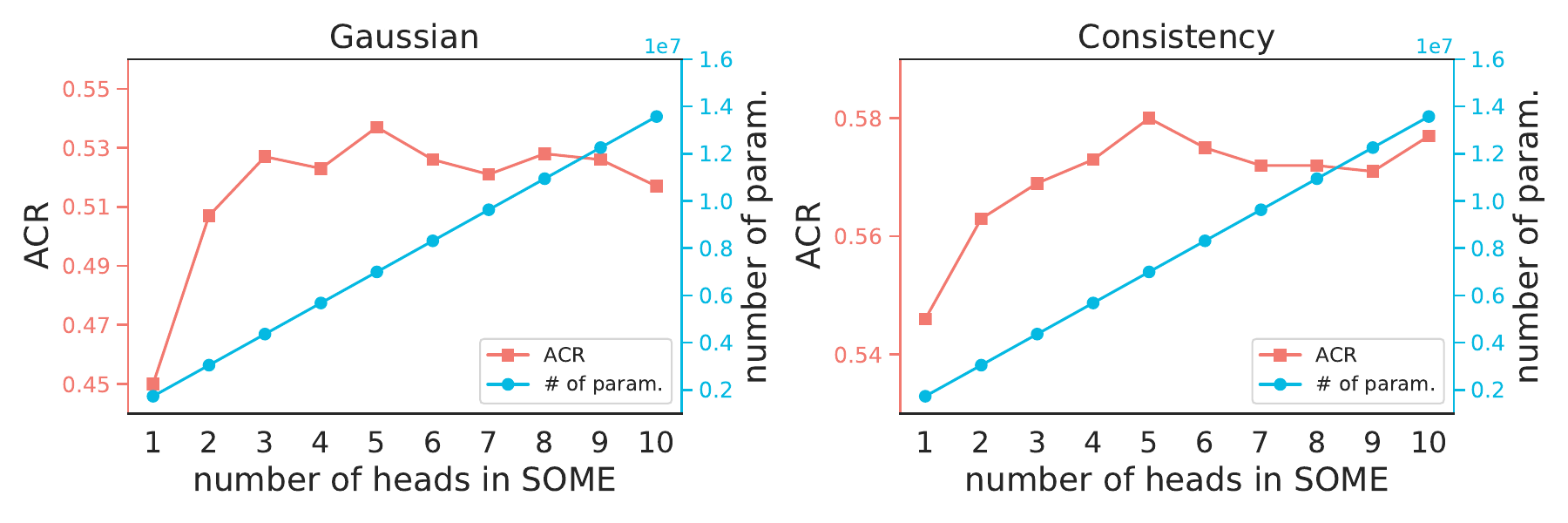}
   \caption{The performance of SOME under varied numbers of heads. Experiments are executed based on ResNet-110 on CIFAR10 with a noise level $\sigma=0.25$.}
   \label{fig:r2.3-varied-heads}
\end{figure}

\begin{figure*}[ht]
  \centering
   \subfigure[Soft and hard thresholding in SPL.]{\includegraphics[width=0.3\linewidth]{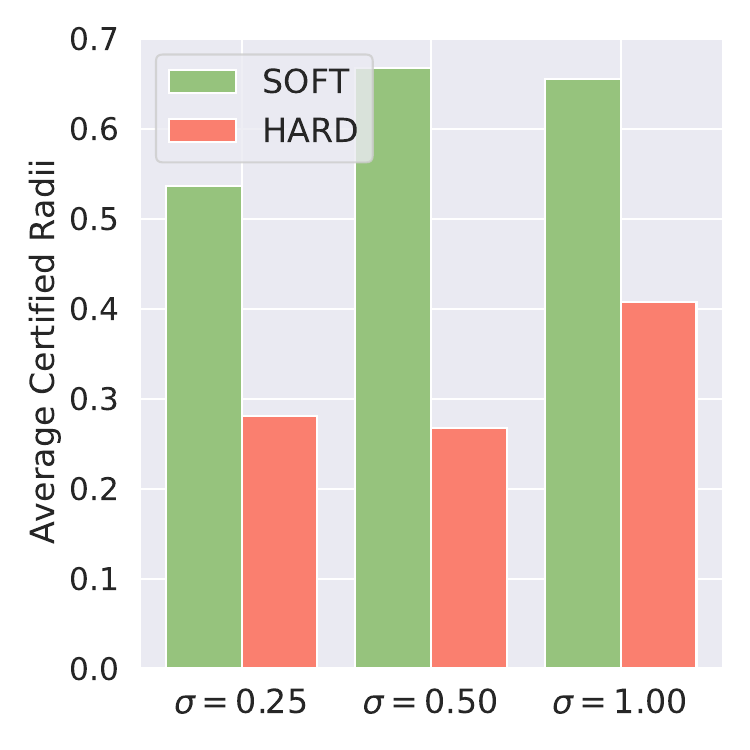}\label{fig:r1.2-ablation-hard}}
   \subfigure[Larger weighting on high-loss end.]{\includegraphics[width=0.3\linewidth]{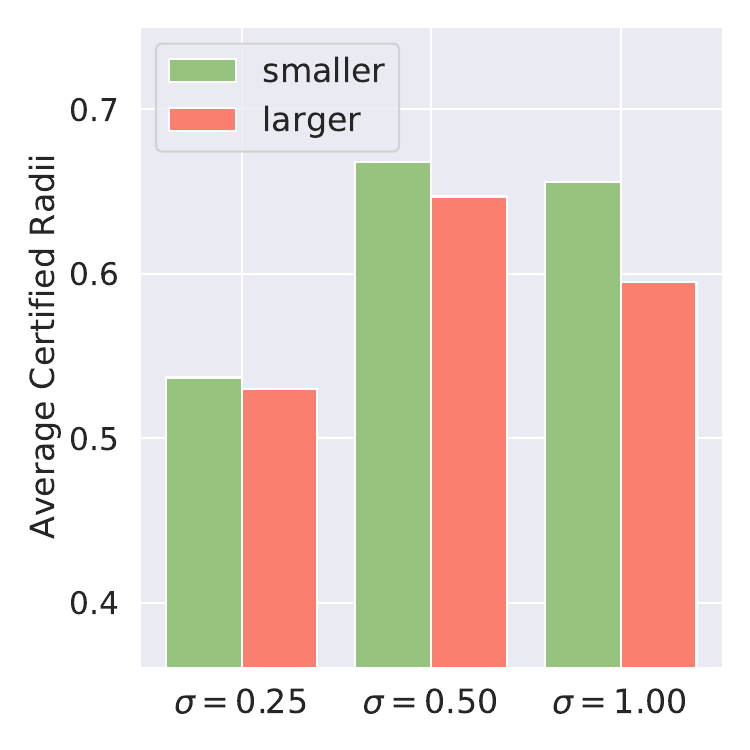}\label{fig:r1.2-ablation-high}}
   \subfigure[Ablations on the circular scheme.]{\includegraphics[width=0.3\linewidth]{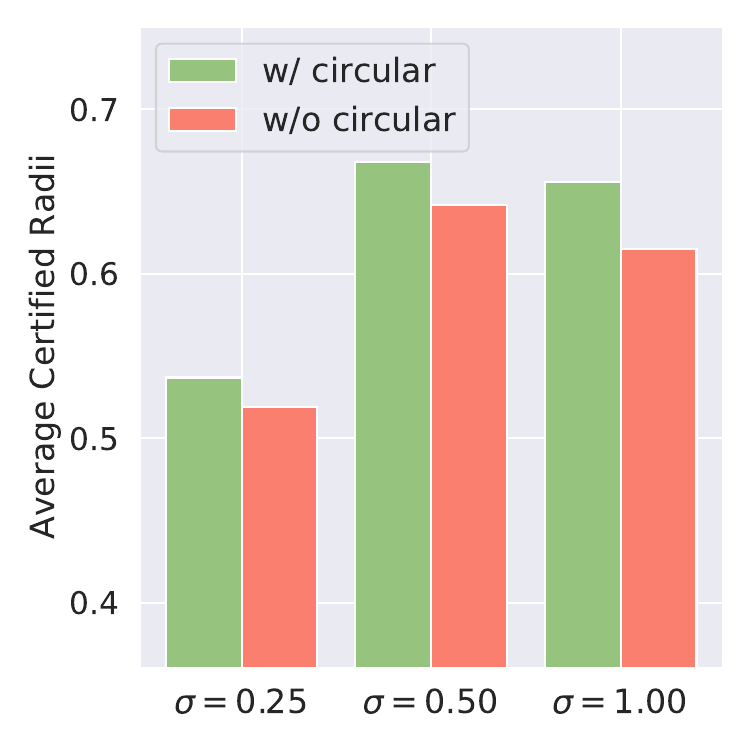}\label{fig:r1.2-ablation-circular}}
   \caption{Ablation studies on the thresholding scheme of SOME. 
   {\bf (a)}: Ablations on the soft and hard thresholding in the self-paced learning of SOME. 
   {\bf (b)}: Ablations on the larger weighting on the high-loss end.
   {\bf (c)}: Ablations on the circular-teaching scheme of SOME.
   All the SOME models are of ResNet110 and are trained with the base method Gaussian~\cite{cohen2019certified} on CIFAR10 under noise levels $\sigma\in\{0.25,0.50,1.00\}$.}
   \label{fig:r1.2-ablation-weights}
\end{figure*}

\subsubsection{On the number of heads in SOME}
\label{app:ablation-number-of-heads}

The number of heads $L$ in SOME balances the trade-off between the certified robustness and the computational cost in such an multi-head structure.
In the Figure \ref{fig:r2.3-varied-heads}, the ACR for certified robustness and the number of parameters of SOME are jointly demonstrated under a variety of different numbers of heads.
As the number of heads increases, the ACR increases and gradually gets stable, indicating that enlarging model capacities could help enhance the certified robustness.
Meanwhile, the increasing model parameters imply huge computational costs with longer training time.
Therefore, the choice on numbers of heads should be considered \emph{w.r.t} the trade-off between performance and computation cost. 
In our experiments, we adopt a 5-head ResNet110 on CIFAR10 and a 3-head ResNet50 on ImageNet, which is able to achieve superior certifiably robust DNNs without  much extra training costs.

\subsection{Empirical discussions on the thresholding scheme}
\label{sec:ablation-thresholding}

In this section, we consider 3 different thresholding schemes in the Self-Paced Learning (SPL) of training SOME models.

\subsubsection{On the hard thresholding for the SPL weights}
The hard and soft thresholding in SPL differs in the assigned weight for those hard samples.
In the {\it hard} thresholding, those hard samples are assigned with zero weights with $\nu_n^k=0$, and only the easy samples are used in training.
In contrast, in SOME, the {\it soft} thresholding is adopted where hard samples are given a small weight $\nu_n^k<1$ (see Eq.\eqref{eq:weights}).
Accordingly, soft thresholding enables that hard samples also contribute to the model training but with smaller weights in the loss function, which could strengthen SPL with performance improvements.

Experiments are conducted where SOME models are trained under the hard thresholding of SPL, shown in the left panel of \cref{fig:r1.2-ablation-weights}.
Under the hard thresholding of the SPL, a number of hard samplers are not considered in training SOME models, leading to a substantial performance drop. Therefore, it is necessary to apply the soft thresholding with weights on hard samples in the SPL of SOME, as explained above.

\subsubsection{On the larger weighting on the high-loss end}
In SOME, easy samples with smaller smoothed losses are assigned with larger weights in the loss function, so that models could learn more from easy samples.
In this section, we consider an opposite setup where larger weights are assigned to hard samples with higher smoothed losses.
In this setup, models pay more attention to hard samples during training.

As shown in the middle panel of \cref{fig:r1.2-ablation-weights}, larger weighting on high-loss end leads to a performance drop, especially on cases of large noises $\sigma=0.50$ and $\sigma=1.00$. 
Therefore, it is proved effective to set larger coefficients for easy samples with smaller smoothed losses, so that the optimization proceeds in a meaningful order from easy samples to hard samples.

\subsubsection{On the self weighting for the SPL weights}
The circular teaching scheme in SOME enables the sample to exchange among these heads, so that each classifier learns from its neighborhood, which is inspired from the celebrated co-teaching framework~\cite{han2018co}.
Ablation studies are executed by cancelling the circular communication in SOME.
In this setup, each head is optimized based on its own easy and hard samples without any available information from other heads.

The right panel of \cref{fig:r1.2-ablation-weights} demonstrates the comparison results and validates that the circular-teaching scheme boosts certified robustness, as the communication among multiple heads enhances the diversities of the classifiers by the diverse sample information  from neighboring heads.

\begin{table*}
  \caption{Runtime comparisons among Ensemble~\cite{horvath2021boosting}, DRT~\cite{yang2021certified} and our SOME. $K$ denotes the number of deployed DNNs.}
  \label{tab:comphensive-comparison-ensemble}
  \centering
  \resizebox{\textwidth}{!}{
  \begin{tabular}{@{}c|l|c|HcHc|cc|Hc@{}}
    \toprule
    \multirow{2}{*}{$\sigma$} & \multirow{2}{*}{Methods} & \multirow{2}{*}{ACR} & \multirow{2}{*}{\# Params.} & \multirow{2}{*}{Base methods$\times K$} & \multirow{2}{*}{Models} & \multirow{2}{*}{GFLOPs$\times K$} & \multicolumn{2}{c|}{Training} & \multicolumn{2}{c}{Certification} \\
    & & & & & & & mem. (MiB) & per epoch (s) & mem. (MiB) & per sample (s) \\
    \midrule
    \multirow{3}{*}{0.25} & Ensemble~\cite{horvath2021boosting}& 0.575 & $-$ & Consistency$\bf\times10$ & & 0.256$\bf\times10$ & 4,197$\bf\times10$ & 43.0$\bf\times10$ & 2,469 & 78.5 \\
    & DRT~\cite{yang2021certified} & 0.551 & $-$ & Gaussian$\bf\times3$ & & 0.256$\bf\times3$ & 21,143 & 21.5$\bf\times3$ + 1055.0 & 22,805 & 24.4 \\
    & SOME ({\bf5-head}) & 0.577 & $-$ & SmoothMix$\bf\times1$ & & 0.594$\bf\times1$ & 9,267 & 461.7 & 2,407 & 12.3 \\
    
    \midrule
    
    \multirow{3}{*}{0.50} & Ensemble~\cite{horvath2021boosting}& 0.754 & $-$ & Consistency$\bf\times10$ & & 0.256$\bf\times10$ & 4,197$\bf\times10$ & 43.4$\bf\times10$ & 2,469 & 78.5  \\
    & DRT~\cite{yang2021certified} & 0.760 & $-$ & SmoothAdv$\bf\times3$ & & 0.256$\bf\times3$ & 21,351 & 21.7$\bf\times3$ +2,192.8 & 22,805 & 24.3  \\
    & SOME ({\bf5-head}) & 0.753 & $-$ & SmoothMix$\bf\times1$ & & 0.594$\bf\times1$ & 9,267 & 460.7 & 2,407 & 12.2  \\
    
    \midrule    
    
    \multirow{3}{*}{1.00} & Ensemble~\cite{horvath2021boosting}& 0.815 & $-$ & SmoothAdv$\bf\times10$ & & 0.256$\bf\times10$ & 4,219$\bf\times10$ & 251.0$\bf\times10$ & 2,469 & 78.3  \\
    & DRT~\cite{yang2021certified} & 0.868 & $-$ & SmoothAdv$\bf\times3$ & & 0.256$\bf\times3$ &  21,351 & 21.5$\bf\times3$+2,178.1 & 22,805 & 24.5  \\
    & SOME ({\bf5-head}) & 0.818 & $-$ & SmoothMix$\bf\times1$ & & 0.594$\bf\times1$ & 9,267 & 464.1 & 2,407 & 12.4  \\
    \bottomrule
  \end{tabular}}
\end{table*}

\subsection{Comparisons with the DRT method}
Another ensemble-based certified defense DRT~\cite{yang2021certified} gives theoretical discussions on the sufficient and necessary conditions for certifiably-robust ensemble models, \textit{i.e.}, diversified gradient and large confidence margin, and accordingly proposes an ensemble training strategy by leveraging the two conditions to train ensemble models.
Despite of the  theoretical analyses, DRT~\cite{yang2021certified} is not computational friendly for its completely different training scheme from other certified defenses.
In existing single-model-based~\cite{cohen2019certified,jeong2020consistency,jeong2021smoothmix,zhai2019macer,salman2019provably} or ensemble-based~\cite{horvath2021boosting} certifiably-robust defenses, models are all trained from scratch.
In contrast, DRT starts from multiple well-pretrained models, \textit{e.g.}, $K$ DNNs trained on CIFAR10 for 150 epochs as~\cite{cohen2019certified}, and particularly, continues to fine-tune these $K$ well-pretrained DNNs for another 150 epochs. The training (pretraining) of multiple DNNs already requires quite a lot computational cost, and yet the fine-tuning of DRT needs much more computational expenses than the overhead computation in the pretraining.
Besides, as the gradient diversity loss term in DRT involves gradients on input images, the resulting GPU memory occupation during training  also inevitably increases.

\Cref{tab:comphensive-comparison-ensemble} gives a comprehensive runtime comparison for SOME with the end-to-end ensemble defense~\cite{horvath2021boosting} and the fine-tuning method DRT~\cite{yang2021certified}.
Because the models and their corresponding settings of~\cite{yang2021certified} are not released, we employ the best ACR values reported in their paper. In~\cite{yang2021certified}, the certification settings are different as they certify the test set every 10 images. 
For a fair comparison with the results released by \cite{yang2021certified}, we re-certify for Horv{\'a}th \textit{et al.}~\cite{horvath2021boosting} and SOME following the same settings in~\cite{yang2021certified}. 
% For Horv{\'a}th \textit{et al.}~\cite{horvath2021boosting}, certifications are re-executed based on the setups with the best results reported in their paper.
In \cref{tab:comphensive-comparison-ensemble}, each experiment is executed on one single GPU of NVIDIA GeForce RTX 3090 (Ampere) with 24GB memory, as our previously used  NVIDIA GeForce RTX 3060 (Ampere, 12GB memory) in \cref{sec:exp-runtime} lacks capacity for DRT~\cite{yang2021certified}. 
All the models are trained on CIFAR10 with a batch size of 256. The pretraining of DRT~\cite{yang2021certified} all requires  4189$\bf\times3$ MiB memory, which is omitted in the table for a more succinct comparison.

In \cref{tab:comphensive-comparison-ensemble}, as a single DNN augmented with 5 heads, SOME achieves competitive or even better ACR over the ensemble of up to 10 individual DNNs~\cite{horvath2021boosting}, with the much saved time expenses in certification.
On the other hand, SOME can still outperform DRT~\cite{yang2021certified} when $\sigma=0.25$.
Notice that such ACR gain is from the  fine-tuning multiple well-pretrained DNNs at a very  high memory occupation and training time-consuming, as shown in \cref{tab:comphensive-comparison-ensemble}.  In this paper, we mainly focus on the efficient  ensemble method trained in an end-to-end way, involving  a circular communication flow along classifiers.
It would also be of  interest in future works to apply SOME on multiple DNNs, either by the training from scratch or by the fine-tuning.

\section{Conclusion}
\label{sec:conclusion}
Based on the ensemble in RS for certified robustness, we propose SOME, which includes (i) an augmented multi-head DNN structure, and (ii) an associated ensemble training strategy circular-teaching.
The multi-head structure seeks for the ensemble via variance reduction with an imposed cosine constraint and greatly alleviates the  heavy computational cost in the existing ensemble methods using multiple DNNs.
The proposed circular-teaching allows communication and information exchange among  heads via the  self-paced learning, which is  modified with the proposed smoothed loss in specific relations to the certified robustness of samples.
In SOME, each head selects the mini-batch samples to teach the training of its neighboring head, forming a circular communication flow for better optimization results and enhanced diversities.
Through extensive experiments, SOME achieves  SOTA results over the existing  ensemble methods at the cost of distinctively less computational expenses. In future, different ensemble mechanisms are worthy to be investigated to further boost certified robustness and efficiency.

The limitation of SOME mainly lies in that the hyper-parameter $\lambda$ needs to be tuned, as it determines the selection of easy and hard samples and affects the procedure of self-paced learning for certified robustness.
Therefore, how to self-adaptively learn $\lambda$ could be further explored to facilitate the application of SOME for various data distributions and networks.

%%%%%%%%%%%%%%%%%%%%%%%%%%%%%%%%%%%%%
%%%%%%%%%%%%%%%%%%%%%%%%%%%%%%%%%%%%%
%%%%%%%%%%%%%%%%%%%%%%%%%%%%%%%%%%%%%

%%%%%%%%% REFERENCES
\bibliographystyle{unsrt}  
\bibliography{reference}

% \begin{comment}
\clearpage

\begin{appendices}
\section{More related works and compared methods}
\label{app:related-works}
The general concept of multi-head structures with DNNs is employed in different fields. 
In the prevailing vision transformer~\cite{dosovitskiy2020image,liu2021swin}, multi-head attention is devised, where each attention head performs scaled dot-products on queries, keys and values, instead of the common convolutions on features.
Attention from multiple heads then presents the model with information from different representation subspaces at different positions to capture dependencies hidden in the sequence of image patches. 
Such revised  attention blocks are  intrinsically different from our multi-head structure in both technical aspects and tackled tasks.
In medical image analysis,  Linmans \textit{et al.}~\cite{linmans2020efficient} considers the out-of-distribution detection for normal lymph node tissue from breast cancer metastases, and deploys a multi-head convolutional neural network for more efficient computations~\cite{lakshminarayanan2017simple}. This work simply pursues the efficiency for the binary classification on very high-resolution images, where no considerations are involved either in the optimization techniques or for the specific tasks. 
In against adversarial attacks, an empirical defense is proposed in~\cite{fang2024towards} by simply using multiple last linear layers, enhancing the network robustness both with and without adversarial training. This method is actually not claimed as an ensemble method, since only a single last linear layer is used for predictions at each forward inference, as  in single-model methods. 

In existing ensemble methods for certified robustness, multiple individual DNNs are employed for ensemble prediction, and each of those individual DNNs in fact simply realizes any existing single-model-based method with RS \cite{horvath2021boosting}. Similarly in our proposed method, each head pertaining a classifier can also be implemented with different methods used in single models. To be more specific, we give details on how the proposed SOME augments the 3 representative  single-model-based certified defenses in each head as done in the experiments.

\noindent\textbf{Gaussian.}
The training of the Gaussian method with RS~\cite{cohen2019certified} simply includes the cross-entropy loss on Gaussian corrupted samples. 
SOME leveraging the Gaussian method in each head is described in \cref{alg:some}, and we exemplify the corresponding optimization problem in the form of expected risk minimization as follows:
\begin{equation}
\label{eq:SPACTE+gaussian}
\mathop{\min}_{g,h_1,\cdots,h_L}\mathop{\mathbb{E}}_{(\mathbf{x},y),\boldsymbol\delta}
\left[\sum_{k=1}^L\nu^{k-1}\mathcal{L}_{\rm ce}\left(f^k(\mathbf{x}+\boldsymbol\delta),y\right)\right]+\mathcal{L}_{\rm cos}.
\end{equation}

\noindent\textbf{Consistency.}
The training of the Consistency method with RS~~\cite{jeong2020consistency} introduces two regularization terms to additionally restrict the predictive consistency in $\mathcal{N}(\mathbf{x},\sigma^2\boldsymbol{I})$.
Given a classifier $f:\mathbb{R}^d\mapsto\mathbb{R}^C$, the optimization objective is formulated as:
\begin{equation}
\label{eq:consistency}
\begin{split}
&\mathop{\min}_{f}\ \mathop{\mathbb{E}}_{(\mathbf{x},y),\boldsymbol\delta}\left[\mathcal{L}_{\rm clf}+\mathcal{L}_{\rm con}\right],\\
&\mathcal{L}_{\rm clf}=\mathcal{L}_{\rm ce}\left(f(\mathbf{x}+\boldsymbol\delta),y\right),\\
&\mathcal{L}_{\rm con}=c_1\cdot\mathrm{KL}\left(f(\mathbf{x}+\boldsymbol\delta)||\hat{f}(\mathbf{x})\right)+c_2\cdot\mathrm{H}\left(\hat{f}(\mathbf{x})\right),
\end{split}
\end{equation}
where $\hat{f}(\mathbf{x})\coloneqq\mathbb{E}_{\boldsymbol\delta} \left[f(\mathbf{x}+\boldsymbol\delta)\right]$.
In \cref{eq:consistency}, aside of the classification loss $\mathcal{L}_{\rm clf}$, the additional loss $\mathcal{L}_{\rm con}$ forces a small KL divergence between the predictions on one noisy sample $f(\mathbf{x}+\boldsymbol\delta)$ and the mean predictions on multiple noisy samples $\mathbb{E}_{\boldsymbol\delta} \left[f(\mathbf{x}+\boldsymbol\delta)\right]$, so as to pursue the predictive consistency on $\mathcal{N}(\mathbf{x},\sigma^2\boldsymbol{I})$ for larger certified radii.
The mean predictions $\hat{f}(\mathbf{x})$ are also restricted by the entropy function $\rm H(\cdot)$.
$c_1$ and $c_2$ are in correspondence with the coefficients $\lambda$ and $\eta$, respectively, in the original paper~\cite{jeong2020consistency}.
Notice that the additional term $\mathcal{L}_{\rm con}$ is sample-wisely determined and thus can be weighted via the SPL strategy in SOME.
The optimization objective of SOME based on Consistency is written as:
\begin{equation}
\label{eq:SPACTE+consistency}
\begin{split}
&\mathop{\min}_{g,h_1,\cdots,h_L}\mathop{\mathbb{E}}_{(\mathbf{x},y),\boldsymbol\delta}
\left[\sum_{k=1}^L\nu^{k-1}\left(\mathcal{L}^k_{\rm clf}+\mathcal{L}^k_{\rm con}\right)\right]+\mathcal{L}_{\rm cos},\\
&\mathcal{L}^k_{\rm clf}=\mathcal{L}_{\rm ce}\left(f^k(\mathbf{x}+\boldsymbol\delta),y\right),\\
&\mathcal{L}^k_{\rm con}=c_1\cdot\mathrm{KL}\left(f^k(\mathbf{x}+\boldsymbol\delta)||\hat{f}^k(\mathbf{x})\right)+c_2\cdot\mathrm{H}\left(\hat{f}^k(\mathbf{x})\right).
\end{split}
\end{equation}

\noindent\textbf{SmoothMix}. The training of the SmoothMix method~\cite{jeong2021smoothmix} follows the idea of adversarial training~\cite{madry2018towards}, where the adversarial examples generated on the smoothed classifier are incorporated into training the base classifier.
Given a classifier $f:\mathbb{R}^d\mapsto\mathbb{R}^C$, the optimization problem of Smoothmix is formulated as:
\begin{equation}
\label{eq:smoothmix}
\begin{split}
&\mathop{\min}_{f}\ \mathop{\mathbb{E}}_{(\mathbf{x},y),\boldsymbol\delta}\left[\mathcal{L}_{\rm clf}+\mathcal{L}_{\rm mix}\right],\\
&\mathcal{L}_{\rm mix}=c_3\cdot\mathrm{KL}\left(f\left(\mathbf{x}^{\rm mix}\right)||y^{\rm mix}\right),\\
&\mathbf{x}^{\rm mix}=(1-c_4)\cdot\mathbf{x}+c_4\cdot\mathbf{\tilde{x}}^{(T)},\\
&y^{\rm mix}=(1-c_4)\cdot\tilde{f}(\mathbf{x})+c_4\cdot\frac{1}{K},c_4\sim\mathcal{U}(\left[0,\frac{1}{2}\right]).
\end{split}
\end{equation}
In \cref{eq:smoothmix}, the additional term $\mathcal{L}_{\rm mix}$ exploits the mix-up between the adversarial examples and clean images, so as to calibrate those over-confident examples via linear interpolation for re-balancing certified radii.
$\mathbf{\tilde{x}}^{(T)}$ indicates the PGD-based $T$-step adversarial examples on the smoothed classifier, and $\tilde{f}(\mathbf{x})$ denotes the soft predictions with softmax on $\mathbb{E}_{\boldsymbol\delta}\left[f(\mathbf{x}+\boldsymbol\delta)\right]$.
$c_3$ and $c_4$ are in correspondence with the coefficients $\eta$ and $\lambda$, respectively, in the original paper~\cite{jeong2021smoothmix}.
Details of generating $\mathbf{\tilde{x}}^{(T)}$ can refer to~\cite{jeong2021smoothmix}.
Similar to the case in Consistency, the SPL strategy in SOME can also be applied to weight the sample-wise loss $\mathcal{L}_{\rm mix}$.
Then, the optimization objective of SOME based on SmoothMix is written as
\begin{equation}
\label{eq:SPACTE+smoothmix}
\begin{split}
&\mathop{\min}_{g,h_1,\cdots,h_L}\mathop{\mathbb{E}}_{(\mathbf{x},y),\boldsymbol\delta}
\left[\sum_{k=1}^L\nu^{k-1}\left(\mathcal{L}^k_{\rm clf}+\mathcal{L}^k_{\rm mix}\right)\right]+\mathcal{L}_{\rm cos},\\
&\mathcal{L}^k_{\rm mix}=c_3\cdot\mathrm{KL}\left(f^k\left(\mathbf{x}^{\rm mix}\right)||y^{\rm mix}\right).
\end{split}
\end{equation}
Notice that here $(\mathbf{x}^{\rm mix},y^{\rm mix})$ is accordingly modified to be generated based on the ensemble of multiple heads.

\section{Experiments on CIFAR10}
\label{app:exp-cifar10}

In this section, setups in relation to results in \cref{tab:comphensive-comparison-cifar10} on CIFAR10 are provided, aside with more empirical results and analysis.
Besides, the training commands for each experiment are all provided in the open-sourced code, where the specified values of the hyper-parameters are presented.

\subsection{Basic setups}
In CIFAR10~\cite{krizhevsky2009learning}, there are 50,000 training images and 10,000 test images of $32\times32\times3$ size categorized into 10 classes.
In all the certifications, if not specified, we follow the same settings in  previous works~\cite{cohen2019certified,horvath2021boosting}, and on CIFAR10 we certify a subset of 500 images by sampling every 20 images from the test set without shuffling.
The related parameters in certification~\cite{cohen2019certified,jeong2020consistency,jeong2021smoothmix,horvath2021boosting,yang2021certified} also follow the common settings: $n_0=100$, $n=100,000$ and $\alpha=0.001$.

To train the ResNet-110~\cite{he2016deep} models on CIFAR10, the stochastic gradient descent optimizer is adopted with Nesterov momentum (weight 0.9, no dampening) with weight decay 0.0001. 
A batch size of 256 is employed.
The initial learning rate is 0.1, which is reduced by a factor of 10 every 50 epochs.

\begin{figure}[t!]
  \centering
   \includegraphics[width=0.6\linewidth]{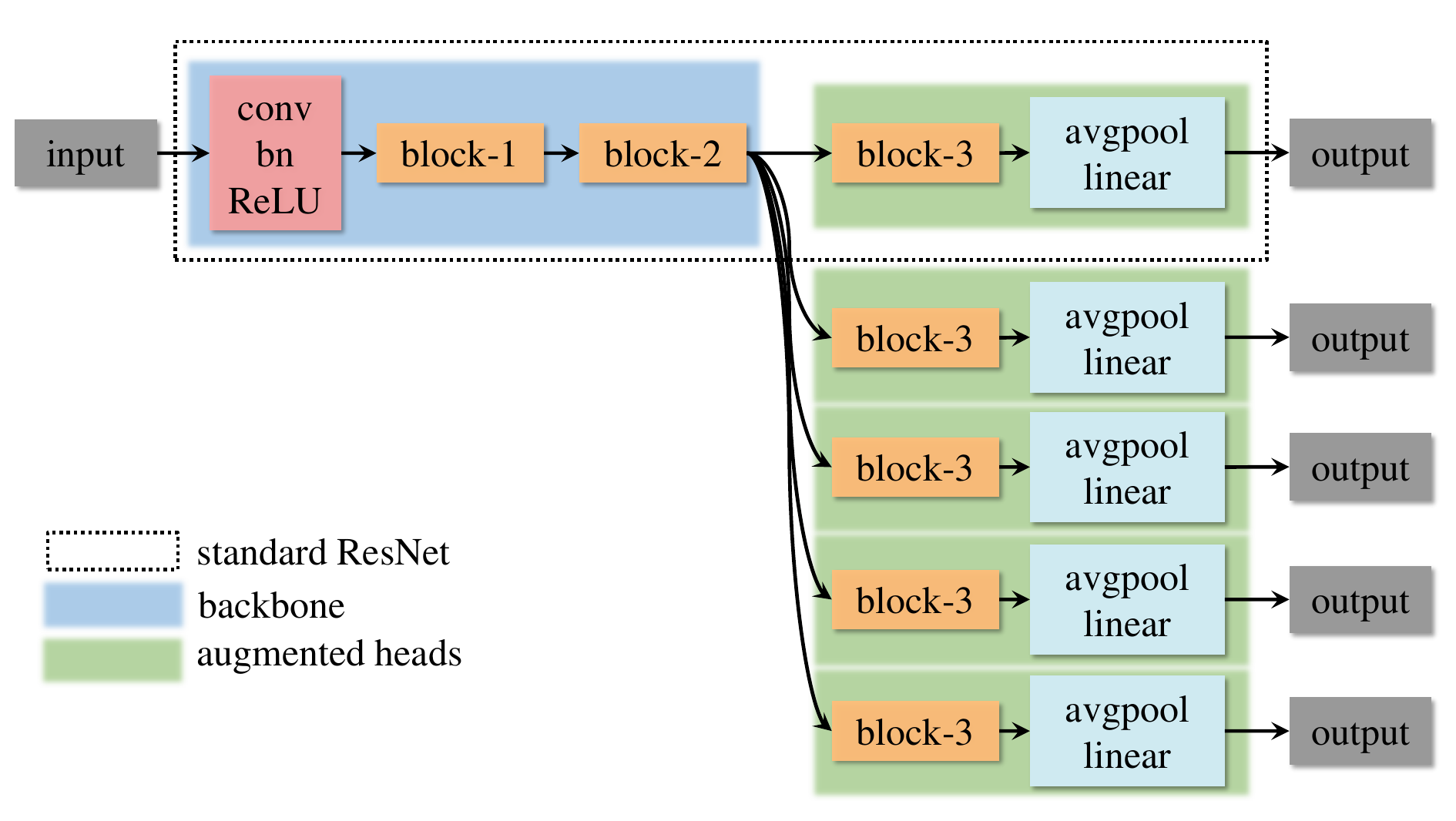}
   \caption{An illustration on where to augment multiple heads in a typical residual network.}
   \label{fig:app-arch}
\end{figure}

\subsection{Settings on the multi-head structure}

We augment multiple heads in the ResNet-110 on CIFAR10.
In all the experiments, we adopt  5 heads, \textit{i.e.}, $L=5$. 
The 5-head network is an appropriate choice from extensive empirical evaluations, as too many heads make the training difficult and too few heads cannot provide enough model capacities.
\Cref{fig:app-arch} gives details on a 5-head DNN.
The dotted black box frames out the typical structure of a standard residual network: a concatenation of a convolution-batchnorm-ReLU layer and 3 residual blocks followed by an average pooling layer and the last linear layer.
The certain location where the heads get augmented is placed right after the second residual block. 
That is, each augmented head consists of the third residual block, the average pooling layer and the linear layer, represented by the green shaded area in \cref{fig:app-arch}.
Accordingly, the backbone, shown by the blue shaded area, contains the convolution-batchnorm-ReLU layer and the first 2 residual blocks.

We now dive into more details on such a 5-head network.
Without loss of generality, we exemplify the deployed ResNet-110.
The 1st, 2nd and 3rd residual blocks contain 34, 37 and 37 convolution layers, respectively.
Thereby the number of augmented layers in one head is only $37+1=38$ (the 3rd block and the last linear layer), accounting for approximately $38/110\approx34\%$ in a 110-layer residual network.
Though, it is worth mentioning that the parameters are not uniformly distributed in a residual network.
The back layers near outputs generally have more parameters than the front layers near inputs due to the increasing number of channels along layers.
Accordingly, each head saves $\approx 66\%$ network layers and $\approx 25\%$ parameters in a 110-layer residual network with in total 1,730,714 parameters. 
Thus, such a multi-head network helps save lots of computations compared with the ensemble of multiple individual DNNs, which has been empirically elaborated in detail from several aspects in \cref{sec:exp-runtime}. 
More empirical evidences on the performance of our multi-head structure augmented in different layers have been illustrated in \cref{fig:ablation-where-to-arg} of \cref{app:ablation-head-location}.

\subsection{Settings related to the SPL coefficients}

In experiments, the number of samples $m$ is used to sample $m$ Gaussian-corrupted images, so as to compute the smoothed loss.
Existing prevailing certified defenses also require multiple samplings during training~\cite{jeong2020consistency,jeong2021smoothmix}.
Following the empirical practice from these methods, $m=2$ is adopted to train SOME models, as too many samplings increase the training time-consuming but do not bring significant performance improvements.

\begin{figure}[t!]
  \centering
  \includegraphics[width=0.6\linewidth]{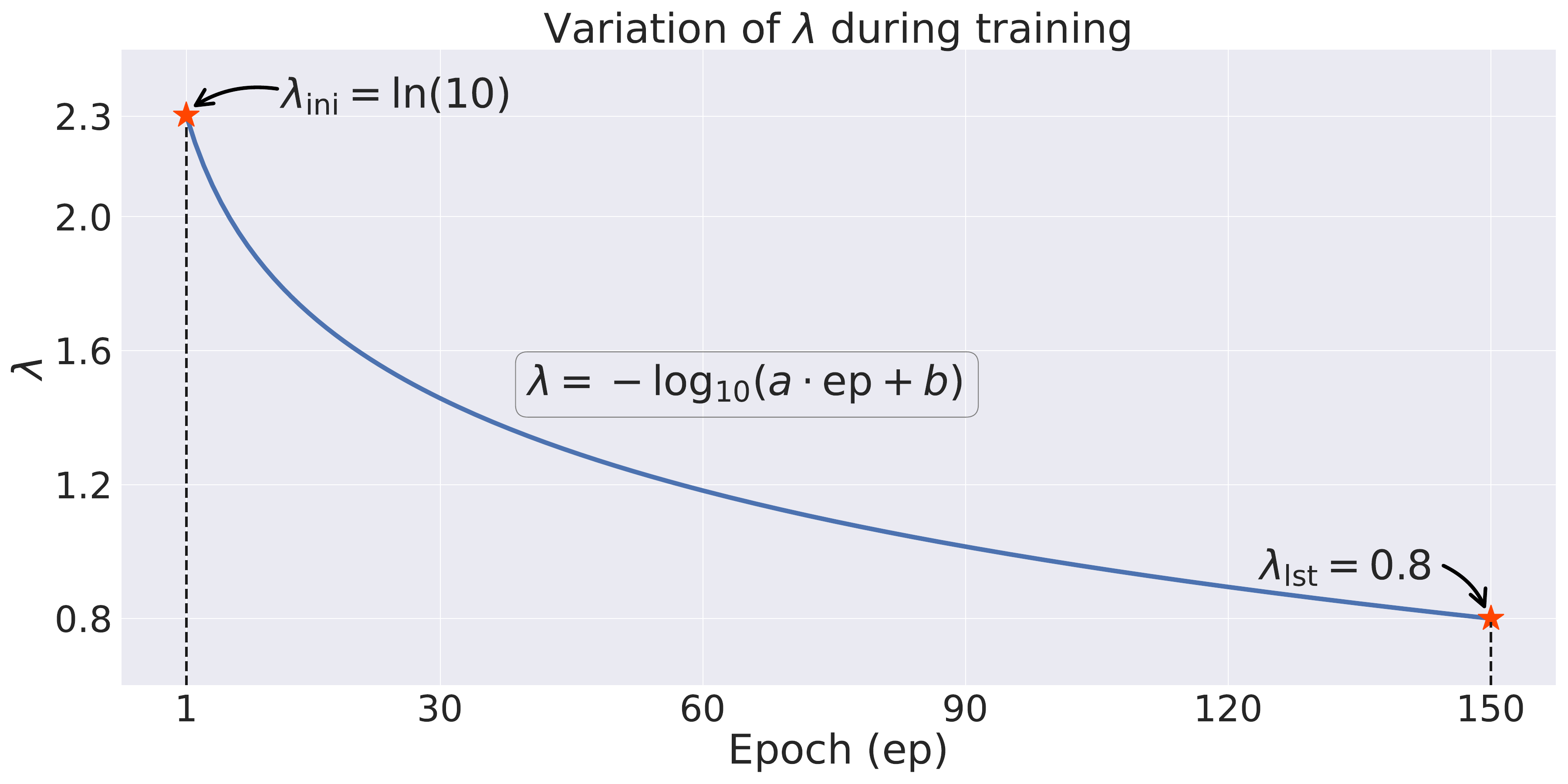}
  \caption{The variation curve of $\lambda$ during training. Given $\lambda_{\rm ini}$ and $\lambda_{\rm lst}$, the coefficients $a$ and $b$ of the curve could be determined, which then specifies the detailed $\lambda$ values at every training epoch.}
  \label{fig:variation_lbd}
\end{figure}

The threshold $\lambda$ in \cref{eq:weights} to control the easy and hard samples is scheduled dynamically during training.
$\lambda$ varies in a logarithmic shape with base 10.
Given the initial value $\lambda_{\rm ini}$ at the 1st epoch and the last value $\lambda_{\rm lst}$ at the last epoch, the values of $\lambda$ at each training epoch can be determined.
We choose the initial value of $\lambda$ as $\lambda_{\rm ini}=\ln{C}$, where $C$ denotes the number of classes, while $\lambda_{\rm lst}$ could be tuned according to practitioners' demands.
An example of the variation of $\lambda$ is shown in \cref{fig:variation_lbd}.

\subsection{Extending SOME to multiple DNNs}
The proposed circular-teaching can also be extended to train multiple individual DNNs.
It can actually be regarded as a special case of SOME that the entire network right after the input layer is augmented.
In this way, these 5 DNNs are not only jointly trained but also enable communications through the circular-teaching. 
In \cref{tab:exp-ablation-q3}, applying circular-teaching to train these DNNs from scratch consistently improves the ACR and the approximated certified accuracy, particularly in the cases of $\sigma\in \{0.50,1.00\}$, achieving distinctive improvements. 
This further verifies that the communication flow between different classifiers in the ensemble  is indeed effective, providing a promising perspective on boosting general ensemble methods. 
Thus, our proposed method is not only an extension on augmenting a single model, but can also be flexibly applied to enhance the training of ensemble methods. 
We can also see that SOME with the more efficient 5-head structure already achieves comparable accuracy with SOME on 5 DNNs. 
Therefore, training multiple DNNs  via circular-teaching shall be recommended when certified robustness is pursued in the foremost priority and sufficient computation budgets are allowed.

\begin{table*}
  \caption{The proposed circular-teaching can be extended to train multiple DNNs from scratch, showing the strongest certified robustness.}
  \label{tab:exp-ablation-q3}
  \centering
  \resizebox{\textwidth}{!}{
  \begin{tabular}{clc|ccccccccccc}
    \toprule
    \multirow{2}{*}{$\sigma$} & \multirow{2}{*}{Methods} & \multirow{2}{*}{\# ensembles} & \multirow{2}{*}{ACR} & \multicolumn{10}{c}{Radius $r$}\\
    & & & & 0.00 & 0.25 & 0.50 & 0.75 & 1.00 & 1.25 & 1.50 & 1.75 & 2.00 & 2.25 \\
    \midrule
    \multirow{3}{*}{0.25} & Gaussian~\cite{cohen2019certified} & 1 DNN & 0.450 & 77.6 & 60.6 &  45.6 & 30.6 & 0.0 & 0.0 & 0.0 & 0.0 & 0.0 & 0.0 \\
    & Ensemble~\cite{horvath2021boosting} & 5 DNNs & 0.530 & 82.2 & 69.4 & 53.8 & 40.6 & 0.0 & 0.0 & 0.0 & 0.0 & 0.0 & 0.0 \\
    \rowcolor{tabgray}\cellcolor{white} & SOME  & 5 DNNs & 0.541 & 83.0 & 70.8 & 53.2 & 42.2 & 0.0 & 0.0 & 0.0 & 0.0 & 0.0 & 0.0  \\
    \rowcolor{tabgray}\cellcolor{white} & SOME  & 5 heads & 0.537 & 83.4 & 71.6 & 54.4 & 40.4 & 0.0 & 0.0 & 0.0 & 0.0 & 0.0 & 0.0  \\
    
    \midrule
    
    \multirow{3}{*}{0.50} & Gaussian~\cite{cohen2019certified} & 1 DNN & 0.535 & 65.8 & 54.2 & 42.2 & 32.4 & 22.0 & 14.8 & 10.8 & 6.6 & 0.0 & 0.0 \\
    & Ensemble~\cite{horvath2021boosting} & 5 DNNs & 0.634 & 68.8 & 60.6 & 47.8 & 39.2 & 28.6 & 20.0 & 13.8 & 8.4 & 0.0 & 0.0 \\
    \rowcolor{tabgray}\cellcolor{white} & SOME & 5 DNNs & 0.676 & 69.6 & 60.8 & 53.4 & 41.4 & 31.4 & 23.0 & 15.6 & 10.2 & 0.0 & 0.0 \\
    \rowcolor{tabgray}\cellcolor{white} & SOME & 5 heads & 0.668 & 69.8 & 61.8 & 50.8 & 42.6 & 31.2 & 21.6 & 15.2 & 9.4 & 0.0 & 0.0 \\
    
    \midrule
    
    \multirow{3}{*}{1.00} & Gaussian~\cite{cohen2019certified} & 1 DNN & 0.532 & 48.0 & 40.0 & 34.4 & 26.6 & 22.0 & 17.2 & 13.8 & 11.0 & 9.0 & 5.8 \\
    & Ensemble~\cite{horvath2021boosting} & 5 DNNs & 0.601 & 49.0 & 43.0 & 36.4 & 29.8 & 24.4 & 19.8 & 16.4 & 12.8 & 11.2 & 9.2 \\
    \rowcolor{tabgray}\cellcolor{white} & SOME & 5 DNNs & 0.675 & 50.0 & 46.4 & 39.6 & 32.6 & 27.4 & 23.6 & 18.4 & 14.8 & 12.6 & 10.8 \\
    \rowcolor{tabgray}\cellcolor{white} & SOME & 5 heads & 0.656 & 50.6 & 45.2 & 37.8 & 31.6 & 27.8 & 22.4 & 18.4 & 14.8 & 12.0 & 10.6 \\
    \bottomrule
  \end{tabular}}
\end{table*}

\section{Experiments on ImageNet}
\label{app:exp-imagenet}

In ImageNet~\cite{deng2009imagenet}, there are 1,287,167 training images and 50,000 validation images categorized into 1,000 classes.
Again, in all the certifications, following the same settings in previous works~\cite{cohen2019certified,jeong2020consistency,jeong2021smoothmix,horvath2021boosting,yang2021certified}, on ImageNet we certify a subset of 500 images by sampling every 100 images from the validation set without shuffling and set $n_0=100$, $n=100,000$ and $\alpha=0.001$.

In experiments of~\cref{tab:comparison-imagenet}, we adopt a 3-head ResNet-50 structure with $L=3$.
During the training of SOME, the sampling number $m$ to calculate the smoothed loss is set as $m=2$.
Besides, the hyper-parameter $\lambda$ in SPL is fixed for experiments on ImageNet.
All the experiments are executed on 4 NVIDIA GeForce RTX 4090 GPUs.

\end{appendices}

\end{document}